\documentclass[conference]{IEEEtran}
\IEEEoverridecommandlockouts
\usepackage{amsmath,amssymb,amsfonts}
\usepackage{algorithmic}
\usepackage{graphicx}
\usepackage{textcomp}
\usepackage{xcolor}
\usepackage{comment}
\usepackage{newtxtext}
\usepackage{pgfplots}
\pgfplotsset{compat=1.18}
\def\BibTeX{{\rm B\kern-.05em{\sc i\kern-.025em b}\kern-.08em
    T\kern-.1667em\lower.7ex\hbox{E}\kern-.125emX}}
\usepackage{acronym}
\acrodef{ML}{machine learning}
\acrodef{FL}{federated learning}
\acrodef{SGD}{stochastic gradient descent}
\acrodef{IID}{independent and identically distributed}
\acrodef{non-IID}{non independent and identically distributed}
\acrodef{RMSE}{root mean square error}
\acrodef{DP}{differential privacy}
\acrodef{KLD}{Kullback–Leibler divergence}
\usepackage{xcolor}
\usepackage{ifthen} 
\newboolean{showcomments}
\setboolean{showcomments}{true}
\ifthenelse{\boolean{showcomments}}
{ \newcommand{\mynote}[3]{
		\fbox{\bfseries\sffamily\scriptsize#1}
		{\small$\blacktriangleright$\textsf{\emph{\color{#3}{#2}}}$\blacktriangleleft$}}
	\newcommand{\zzz}[1]{{\setlength{\fboxsep}{2pt}\fcolorbox{black}{yellow}{\textsf{\emph{#1}}}}\xspace}}
{ \newcommand{\mynote}[3]{}
	\newcommand{\zzz}[1]{}}

\usepackage{tikz}
\usepackage{pgfplots}
\usepackage{pgfplotstable}
\usepackage[eulergreek]{sansmath}

\usepackage{amsmath,amssymb,amsfonts}
\usepackage{graphicx}
\usepackage{textcomp}
\usepackage{xcolor}
\usepackage{bbm}
\usepackage{mathtools}
\usepackage{caption}
\usepackage{subcaption}
\usepackage{diagbox}
\usepackage{array}
\usepackage{amsmath}
\usepackage{amssymb}
\usepackage{amsfonts}
\usepackage{amsthm}
\usepackage{mathtools}

\usepackage{breqn}
\usepackage{enumitem}
\usepackage{physics}
\usepackage{thm-restate}
\usepackage{url}
\theoremstyle{definition}
\newtheorem{definition}{Definition}[section]
\theoremstyle{remark}
\newtheorem{remark}{Remark}
\newtheorem{theorem}{Theorem}[section]
\newtheorem{corollary}[theorem]{Corollary}

\usepackage{comment}
\usepackage{nicefrac}
\usepackage[abbreviations]{foreign}
\usepackage[algo2e,ruled,lined,boxed,commentsnumbered, noend, linesnumbered, procnumbered]{algorithm2e}
\usepackage{cleveref}
\allowdisplaybreaks
\newcommand{\sys}{\textsc{FedPriKL}\xspace}
\newcommand{\kld}{\ensuremath{D_{\operatorname{KL}}}\xspace}
\newtheorem{fact}[theorem]{Fact}
\newcommand{\fulltrust}{Trusted\xspace}
\newcommand{\aggregator}{TAgg\xspace}
\newcommand{\distributed}{Dist\xspace}
\crefname{fact}{Fact}{Facts}
\Crefname{fact}{Fact}{Facts}

\renewcommand{\paragraph}[1]{\noindent\textit{#1}}
\begin{document}

\title{Federated and differentially private estimation of KL divergence
}

\author{\IEEEauthorblockN{Sayan Biswas}
\IEEEauthorblockA{
\textit{EPFL}\\
Lausanne, Switzerland \\
sayan.biswas@epfl.ch}
\and
\IEEEauthorblockN{Graham Cormode}
\IEEEauthorblockA{
\textit{University of Oxford}\\
Oxford, UK \\
graham.cormode@cs.ox.ac.uk}
\and
\IEEEauthorblockN{Carsten Maple}
\IEEEauthorblockA{
\textit{University of Warwick}\\
Coventry, UK \\
cm@warwick.ac.uk}
\and
\IEEEauthorblockN{Mary Scott}
\IEEEauthorblockA{
\textit{University of Oxford}\\
Oxford, UK \\
mary.scott@ndm.ox.ac.uk}
}

\maketitle

\begin{abstract}
Measuring distribution drifts is a key task in managing distributed, sensitive data, as it underpins a wide range of federated learning and analytics applications. 
In many practical settings, however, directly sharing such information is either undesirable (\eg due to privacy concerns) or infeasible (\eg due to high communication costs). 
In this work, we present \sys, a novel method for estimating the KL divergence of data across federated computational models under differential privacy (DP) guarantees. 
We establish its theoretical properties, showing that \sys is batch-wise unbiased with low and bounded sensitivity and variance, thereby ensuring strong utility under DP. 
In addition, we present an empirical study that explores parameter choices to optimize accuracy while minimizing communication overhead. 
Our experiments demonstrate that \sys achieves accuracy comparable to a similar sampling-based non-private estimator, while delivering greater stability and accuracy than a baseline variant in which users perturb their data locally to obtain privacy guarantees.
\end{abstract}


\section{Introduction}\label{sec:intro}
Modern applications in data analysis and machine learning work with high-dimensional data to support inferences and provide recommendations~\cite{McCarthy04, Najafabadi15}.
Increasingly, the data to support these tasks comes from individuals who hold their data on personal devices such as smartphones and wearables.
In the federated model of computation~\cite{Kairouz21Survey, Bharadwaj22}, this data remains on the users' devices, which collaborate and cooperate to build accurate models by performing computations and aggregations on their locally held information (\eg, training and fine-tuning small-scale models).

A key primitive needed is the ability to compare the distribution of data held by these clients with a reference distribution. For instance, a platform or a service provider would like to know whether the overall behavior of the data is consistent over time for deploying the best-fitting and most relevant model.
In cases where the data distribution has changed, it may be necessary to trigger model rebuilding or fine-tuning, whereas if there is no change the current model can continue to be used.
Hence, we seek ways to measure the similarity of distributions within the federated setting, where strong privacy and security guarantees must be maintained~\cite{Abadi16, Bonawitz17}, while also ensuring that the computation and communication overheads are bounded.

Among ways to compare probability distributions, the Kullback-Leibler (KL) divergence~\cite{Kullback51}  is a standard information-theoretic measure widely used in statistics and machine learning.
Given two distributions $P_1$ and $P_2$ defined over the same domain, the KL divergence between them captures the information-theoretic difference of $P_1$ from $P_2$, sometimes expressed as the amount of ``surprise'' or ``information gain'' in seeing samples from $P_1$ when expecting $P_2$.
If we have a complete representation of $P_1$ and $P_2$ to work with, then the KL divergence can be computed directly in its closed form.
%However, in the modern-day applications of such federated settings, the test distribution is defined implicitly, based on the empirical frequency of the samples held by a distributed set of clients, and, typically, the domains of these client-side distributions are infeasibly large to adapt a full-sharing-based approach.
%In this paper, we develop a sampling-based method for estimating the KL divergence between distributions under a federated framework with formal privacy guarantees and in a communication-efficient manner. 
However, in federated settings catering to modern-day applications, the test distribution is often defined implicitly by the empirical frequencies of samples held across a distributed set of clients. Typically, the domains of these client-side distributions are too large and the data too sensitive to permit a full-sharing approach. 
A trivial solution is for a central server to collect samples from all clients together in a single location, and to evaluate the KL divergence itself.
Although this method gives an accurate answer, it fails to consider privacy or efficiency.
Furthermore, the parameter space of the clients' data can be extremely large, particularly in the setting of modern-day machine learning.
Thus, this initial \emph{full-sharing} method fails to meet the practical requirements of the scenario.
The cost overhead will be high: communicating all the samples or their empirical frequencies can consume significant bandwidth, since the number of data points or the domain size can be very large.
Moreover, this approach raises privacy concerns for the participants since it  reveals the samples held by each client directly to the server.
Variations on this paradigm which have clients add noise to the histogram of their item frequencies also fail due to the high domain size involved.
Instead, we seek solutions that reduce the overhead, and provide provable privacy guarantees under various models of trust.

%In this paper, we develop a sampling-based method for estimating the KL divergence between such distributions under a federated framework that provides formal privacy guarantees while remaining communication-efficient.
In this paper, we study the problem of estimating the KL divergence between a public reference distribution and an empirical distribution held by federated clients.
Our solution satisfies formal differential privacy constraints and avoids communicating full histograms over the data domain. 
Our approach is to reduce the total communication cost by sampling a subset of clients and items to participate in building a randomized estimator, and to ensure that \ac{DP} is satisfied on the output.
Sampling alone is insufficient to solve our problem since it still reveals the sensitive information of the clients included in the sample~\cite{DworkRoth14}.
%For \ac{DP}, although techniques are known for many problems, the case of KL divergence has not been analyzed previously. 
Furthermore, via the \emph{shift invariance theorem}, Guha et al.~\cite{Guha07}  showed that it is not possible to obtain even a non-private sketch of KL divergence between two unknown distributions in space smaller than that of the domain of the distributions. 
Our core technical ingredient is an unbiased Monte Carlo estimator for the KL divergence, based on sampled likelihood ratios $(P(x)/\Pi(x))$, which allows us to build the first private estimator for this task. 
%However, the possibility of a sampling-based approach to \emph{estimate} the KL divergence between an unknown distribution and a reference distribution has not been studied before.
%\sayan{@Graham: pls check the last part of this paragraph -- hope our claim's not being too bold?}
We present \sys, a sampling-based protocol that queries only a small number of points drawn from the reference distribution and uses secure aggregation to estimate the corresponding probability masses in the private federated distribution. 
This sidesteps the impossibility result by restricting attention to the
%practically relevant 
case of comparing a private population to a known benchmark.

Specifically, in this work, we consider a horizontal federated learning environment consisting of a set of clients that interact with a coordinating server to compute a function of interest. 
%Let $P \in \mathcal{P}_{\mathcal{X}}$ denote the (private) distribution of the sensitive data held collectively by the clients, where each client’s data lies in the domain $\mathcal{X}$. 
The primary research question we solve is to design an estimator of the KL divergence between the distribution of the clients’ private data and a reference distribution representing public knowledge in a fully federated and privacy-preserving manner without needing to share the probability mass for each element in the domain.
%$\mathcal{X}$. 
For simplicity, we assume that all clients independently and honestly follow the protocol (\ie, there are no adversarial clients and no collusion). 
To this end, we first propose a sampling-based unbiased estimator of KL divergence, establish its key utility-related properties, and then leverage it to develop a communication-efficient method for \emph{FED}erated and \emph{PRI}vate estimation of \emph{KL} divergence (\sys).
To our knowledge, \sys is the first communication-efficient federated estimator of $D_{KL}(\Pi \| P)$
between a public reference and a private empirical population that provides formal example-level DP.
In our trust model, the trusted aggregator observes only securely aggregated statistics for the sampled points; confidentiality with respect to this aggregator is supplied by the trust assumption, or by its TEE/SMC realization, rather than by differential privacy.
%To our knowledge, \sys is the first method to enable an efficient and accurate estimation of KL divergence with formal \ac{DP} guarantees.
%Our contributions are as follows. First, we formalize federated KL-divergence estimation under example-level differential privacy, including the adjacency notion and trust assumptions needed for private estimation from distributed empirical data. Second, we derive and analyze the \sys estimator, proving unbiasedness and giving explicit sensitivity bounds that permit calibrated Gaussian noise addition in the trusted-aggregator model. The analysis also characterizes how the estimator variance depends on the sampling budget, privacy parameters, reference-distribution support, and the variance-control parameter (\lambda), thereby making the privacy–accuracy–communication tradeoff explicit. Third, we separate the effect of where privacy noise is introduced by comparing \sys with a natural locally perturbed baseline; this highlights why adding noise after aggregation preserves utility for this nonlinear divergence-estimation task. Finally, we evaluate the method on federated FEMNIST data, showing that \sys attains accuracy close to the corresponding non-private sampling estimator while requiring each participating client to communicate only counts for a small number of sampled domain elements.

Our key contributions are as follows: 
\begin{itemize}
    \item We formalize the problem of federated computation of KL divergence with privacy.
    %, and describe three different computational models, based on differing levels of trust.
    \item
    We propose \sys, a novel probabilistic estimator for KL divergence based on sampling, and provide a comprehensive theoretical analysis that formally characterizes its accuracy. By analyzing the sensitivity of the proposed estimator, we demonstrate how example-level differential privacy guarantees can be achieved through carefully designed noise addition within the federated model, and we provide a theoretical characterization of the optimal parameters to maximize the utility of \sys.
    \item 
    We present an experimental study of our methods applied to real data, and compare the impact of varying different parameters such as the rate of client sampling and the level of privacy provided.
    We see that \sys can obtain similar accuracy to a non-private estimator, while providing a strong formal privacy guarantee. 
\end{itemize}

%In what follows, \Cref{sec:related} describes related work, and we present preliminaries in \Cref{sec:prelim}.
%Our main algorithms and their properties are presented in \Cref{sec:PRIEST}.
%, while further theoretical analysis is shown in \Cref{sec:analysis}.
%Our experiments are in \Cref{sec:expts}, and we make concluding remarks in \Cref{sec:concs}.

\section{Related Work}\label{sec:related}
The federated computation model was introduced in its current form by the \ac{ML} community, with a focus on training \ac{ML} models over distributed data~\cite{McMahan17}.
%Since then, work has remained focused on this \ac{FL}~\cite{Kairouz21Survey}, but also broadened to consider a wider range of computational tasks, under the banner of ``federated analytics'' or ``federated computation''~\cite{Bharadwaj22}.
Since then, research has largely remained focused on this \ac{FL} paradigm~\cite{Kairouz21Survey}, while gradually expanding to encompass a wider range of  tasks, under the banner of ``federated analytics'' or ``federated computation''~\cite{Bharadwaj22}.

In this paper, we focus on designing randomized algorithms that can operate in the federated setting to estimate the similarity of datasets.
This is a task in support of \ac{FL}, where identifying when a distribution has changed can be important in determining when to improve an existing model via retraining or fine-tuning.
Our approach is informed by prior work which has sought to build compact ``sketches'' for information divergences.
We develop a mechanism to privately compare different sets of data across a period of time.
It is not possible to assume both distributions are unknown, because the Shift Invariance Theorem~\cite{Guha07} means that the distance between them is not estimable in this sketching model.
We navigate this issue by instead comparing one publicly known benchmark distribution with a private one, described by a large group of clients involved in an \ac{FL} environment.

Our model of privacy adopts \acf{DP}, a widely-used statistical approach that introduces carefully calibrated randomness into query answers to mask the footprint of any individual in the data~\cite{DworkDP1,DworkRoth14}.
\ac{DP} has been applied to a wide range of tasks, from gathering basic statistics such as counts and histograms~\cite{DworkRoth14} to the optimization of high-dimensional \ac{ML} models~\cite{Abadi16}.
For distance estimation, results on the privacy properties of the Johnson-Lindenstrauss lemma mean that Euclidean distance can be estimated accurately~\cite{Kenthapadi13, Blocki12}.
However, apart from this, the study of \ac{DP} for distance estimation in the federated model has been limited.

There are numerous metrics in the literature for (non-private) distance estimation.
The Fr\'{e}chet distance~\cite{Hou23} has been used in the generative adversarial network (GAN) literature to evaluate the distance between real datasets, but it can only be written in closed form if both distributions are Gaussian.
The H\"{o}lder divergence~\cite{Nielsen17} can only be proper (zero for two identical distributions) given a specific property between conjugate exponents.
The Earth-Mover distance~\cite{Andoni09} produces accurate representations in all spaces except $\ell_1$, causing problems for Laplace distributions.
%Meanwhile, 
The Kullback-Leibler (KL) divergence~\cite{Kullback51} 
%is a standard notion of distance that does not have any of the above restrictions.
%
%KL divergence [7] 
provides a standard information-theoretic comparison that is particularly natural in our setting because one distribution is a fixed public reference. 
In what follows, we make the KL divergence the focus of our study.
%In what folle focus on estimating $D_{KL}(\Pi\|P)$, %while explicitly accounting for the support conditions required for this direction of KL divergence.

%We base our approach on sampling. 
Prior work has studied general approaches to private estimation using Monte Carlo samplers in order to  
%Prior work has sought to approximate the KL divergence outside of privacy constraints. 
%There are a number of ways in which the KL divergence can be approximated.
%Monte Carlo sampling
draw independent and identically distributed (i.i.d.) samples from the unknown distribution~\cite{Heikkila19}; a good estimator is unbiased and has low variance.
However, our focus is on defining a sampler that can be applied for KL divergence in the federated model. 
%Other techniques can be applied when the data is drawn from a parametric distribution, such as Gaussian mixture models~\cite{Hershey07}.
%However, we are not aware of any prior work that has studied KL divergence under DP, or in the federated model.

Other research has considered different models of secure and private computation, trading off the level of trust in different distributed  participants, and removing reliance on a single centralized trusted entity. 
%The problem can be approached from a decentralized perspective: this is much more realistic than with a central server because of data sharing restrictions.
For instance, work has sought to 
aggregate data for learning without compromising on privacy, via secure multi-party computation (SMC) combined with Gaussian noise scaling to protect all data~\cite{Heikkila17}.
Similarly, in the DPrio algorithm~\cite{Keeler23}, clients generate noise and secret-share it with the servers, who then select a small number of clients' noise to add.
Additional steps to protect from malicious clients include zero-knowledge proofs, secret-shared non-interactive proofs~\cite{Corrigan17} and Boolean secret sharing~\cite{Addanki22}.
%This improves the data utility of previous methods and ensures DP guarantees and robustness against malicious clients.

\begin{table}
    \centering
\begin{tabular}{l|l}
Symbol & Meaning \\\hline
$D_{\operatorname{KL}}$ & KL Divergence (Defn \ref{def:KLDiv}) \\
$\varepsilon, \delta$ & Differential privacy parameters \\
$\Delta(f)$ & Sensitivity of function $f$ (Defn \ref{def:sensitivity}) \\
$\psi(h)$ & Mapping from histogram to probability distribution \\
$\eta_{\varepsilon, \delta}$ & (Gaussian) noise added for differential privacy \\
$n$ & Number of participating clients \\
$m$ & Number of samples used to make estimator \\
$T$ & Number of batches for estimator \\
$\phi$ & KL divergence estimation function (Section~\ref{sec:MC}) \\
$P$, $\Pi$ & Data distributions: public ($\Pi$) and private ($P$) \\
$\pi_\min$ & $\min_{x\in\mathcal{X}}\Pi(x)$\\
$r(x_i)$ & Odds ratio of element $x_i$, as $P(x_i)/\Pi(x_i)$ \\
$\tau$ & Clipping threshold for small/zero probabilities

\end{tabular}

    \caption{Summary of important notation}
    \label{tab:notation}
\end{table}

\section{Preliminaries and Background}\label{sec:prelim}

We begin with key concepts from statistics and privacy.  \Cref{tab:notation} lists the main notation needed. 
%\sayan{needs to be reviewed}
\subsection{Probability Distributions \& Divergences}

We consider probability distributions defined empirically by observations of data.
Assume that each data point is drawn from a discrete finite domain $\mathcal{X}$, and write
$\mathcal{P}_{\mathcal{X}}$ to be the space of probability distributions on $\mathcal{X}$.
Given a finite (multi)set of data points $D$, %of size $m$, 
write $D(x)$ for the frequency of $x$ in $D$.
Let $P_D$ be the associated empirical probability distribution, \ie $P_D(x) = \frac{D(x)}{|D|}$.
Given two probability distributions, $P_1$ and $P_2$, it is natural to want to measure the similarity between them.
%The behavior of $P_1$ and $P_2$ in comparison to each other gives useful information, for example the performance of a new model in comparison to an existing model.
There are many commonly used statistical divergence measures.
%For instance, the Total Variation Distance is given by $D_{TV} (P_{1}, P_{2}) = \frac{1}{2} \sum_{x \in \mathcal{X}} |P_{1}(x) - P_{2}(x)|$.
In this paper, our focus will be on the KL divergence between $P_1$ and $P_2$.

\begin{definition}\label{def:KLDiv}
For discrete probability distributions $P_{1}, P_{2} \in \mathcal{P}_{\mathcal{X}}$, the \emph{Kullback-Leibler (KL) divergence} between $P_{1}$ and $P_{2}$, denoted by $D_{\operatorname{KL}} \left(P_{1} \| P_{2} \right)$, is given by:
\begin{align}
\textstyle
 D_{\operatorname{KL}} \left(P_{1} \| P_{2} \right) = \sum_{x \in \mathcal{X}} P_{1}(x) \ln{\left( \frac{P_{1}(x)}{P_{2}(x)} \right)}.
\nonumber
\end{align}
\end{definition}

 Its asymmetry is useful in our setting: we will estimate $D_{KL}
(\Pi \| P)$, measuring departure of the private population $P$ from a fixed public reference $\Pi$. 
As usual for this direction of KL, finite divergence requires $P(x)>0$ wherever $\Pi(x)>0$. 

%The widespread popularity and many applications of KL divergence 
% make it a first choice for comparing probability distributions.

\subsection{Privacy standards}

%\sayan{provide some short intro/build-up}
%For privacy, we adopt the textbook \ac{DP} definition: 

\begin{definition}[Differential privacy~\cite{DworkDP1, DworkRoth14}]\label{def:DP}
%For any input space $\mathcal{X}_{\operatorname{in}}$ and output space $\mathcal{X}_{\operatorname{out}}$, a randomized algorithm $\mathcal{M}$ operating by mapping any dataset with elements from $\mathcal{X}_{\operatorname{in}}$ to an element in $\mathcal{X}_{\operatorname{out}}$ %(\ie  $\mathcal{M}:\mathcal{X}_{\operatorname{in}}^{d}\mapsto \mathcal{X}_{\operatorname{out}}$ for some $d\in \mathbb{N}$) 
%satisfies $(\varepsilon, \delta)$-\ac{DP} if for any two adjacent datasets $D$ and $D'$ (\ie $D$ and $D'$ differing in at most one element), and any measurable $O\subseteq \mathcal{X}_{\operatorname{out}}$, we have $\Pr[\mathcal{M}(D) \in O] \leq e^\varepsilon \Pr[\mathcal{M}(D') \in O] + \delta$.
%\[
%\Pr[\mathcal{M}(D) \in O] \leq e^\varepsilon \Pr[\mathcal{M}(D') \in O] + \delta.
%\]

For any input space $\mathcal{X}_{\operatorname{in}}$ and output space $\mathcal{X}_{\operatorname{out}}$, a randomized algorithm $\mathcal{M}$ that maps any dataset whose entries belong to $\mathcal{X}_{\operatorname{in}}$ to an element in $\mathcal{X}_{\operatorname{out}}$ satisfies \emph{$(\varepsilon,\delta)$-differential privacy (DP)} if, for any two adjacent datasets $D$ and $D'$ (\ie datasets differing in at most one element), and for any measurable $O \subseteq \mathcal{X}_{\operatorname{out}}$, we have $\Pr[\mathcal{M}(D) \in O] \leq e^{\varepsilon} \Pr[\mathcal{M}(D') \in O] + \delta$.
%\[
%\Pr[\mathcal{M}(D) \in O] \leq e^{\varepsilon} \Pr[\mathcal{M}(D') \in O] + \delta.
%\]
\end{definition}

%We note that if, in the above definition, 
When the adjacency relation is defined element-wise on $\mathcal{X}_{\operatorname{in}}$, we obtain  the notion of ``example-level  \ac{DP}''~\cite{FLP}.
We will apply the \ac{DP} definition to outputs that are histograms.
For a finite domain $\mathcal{X}$, let $\mathcal{H}_{\mathcal{X}} = \{ h \ \vert \ h \colon \ \mathcal{X} \mapsto \mathbb{Z}_{\geq 0} \}$ be the space of all datasets (represented as histograms) whose entries belong to $\mathcal{X}$, and let $\mathcal{P}_{\mathcal{X}}$ be the space of all probability distributions on $\mathcal{X}$.
Let the \emph{cardinality} of any $h \in \mathcal{H}_{\mathcal{X}}$ be defined as $|h| \ =\sum_{x \in \mathcal{X}} h(x)$, and let $\psi(h)$ be the empirical distribution on $\mathcal{X}$ that we obtain from $h$.
That is, $\psi \colon \ \mathcal{H}_{\mathcal{X}} \mapsto [0,1]^{|\mathcal{X}|}$ so that
\centerline{$
\psi(h) = \left\{ \psi(h)_x \colon\, \psi(h)_x = {h(x)} / {|h|} \quad \forall x \in \mathcal{X} \right\}.
$}

\begin{definition}[Adjacent histograms]\label{def:adj_hist}
We call histograms %(\ie, datasets) 
$h_{1}, h_{2} \in \mathcal{H}_{\mathcal{X}}$ \emph{adjacent}, denoted $h_{1} \sim h_{2}$, if $|h_{1}|$ and $|h_{2}|$ are finite, and $\exists x_{1}, x_{2} \in \mathcal{X}$, $x_{1} \neq x_{2}$ s.t.: \\
(1) $h_{1}(x_{1}) - h_{2}(x_{1}) = 1$, \\
(2) $h_{1}(x_{2}) - h_{2}(x_{2}) = -1$, and \\
(3) $h_{1}(x) = h_{2}(x) \,\forall\, x \in \mathcal{X} \setminus\{x_{1}, x_{2}\}$.
%\begin{enumerate}[label=$\roman*$.]
%    \item $|h_{1}|$ and $|h_{2}|$ are finite
%    \item there are $x_{1}, x_{2} \in \mathcal{X}$, $x_{1} \neq x_{2}$ such that: \(a.\) $h_{1}(x_{1}) - h_{2}(x_{1}) = 1$, \(b.\) $h_{1}(x_{2}) - h_{2}(x_{2}) = -1$, and \(c.\) $h_{1}(x) = h_{2}(x) \,\forall\, x \in \mathcal{X} \setminus\{x_{1}, x_{2}\}$.
    %\begin{enumerate}[label=\alph*.]
    %    \item $h_{1}(x_{1}) - h_{2}(x_{1}) = 1$
    %    \item $h_{1}(x_{2}) - h_{2}(x_{2}) = -1$
    %    \item $h_{1}(x) = h_{2}(x) \,\forall\, x \in \mathcal{X} \setminus\{x_{1}, x_{2}\}$.
    %\end{enumerate}
%\end{enumerate}
\end{definition}

\Cref{def:adj_hist} describes dataset pairs that have the same cardinality and differ in the count of exactly one attribute by one.
In other words, we call two histograms $h_{1}$ and $h_{2}$ adjacent if they correspond to a pair of datasets with exactly one item substituted for another, but are otherwise identical.
%exactly two values they hold are swapped and the count of every other $x \in \mathcal{X}$ is the same in $h_{1}$ and $h_{2}$.
This gives the notion of ``example-level differential privacy''.

\begin{definition}[Adjacent distributions]\label{def:adj_prob}
We call distributions $P_{1}, P_{2} \in \mathcal{P}_{\mathcal{X}}$ \emph{adjacent}, denoted $P_{1} \sim P_{2}$, if there exist $h_{1}, h_{2} \in \mathcal{H}_{\mathcal{X}}$ such that
$h_{1} \sim h_{2}$ with $P_{1} = \psi(h_{1})$ and $P_{2} = \psi(h_{2})$.
\end{definition}

\begin{definition}[Sensitivity]\label{def:sensitivity}
For any query $f \colon \mathcal{P}_{\mathcal{X}}\mapsto \mathbb{R}$ on the space of distributions, the \emph{sensitivity of $f$} is:
\[
\Delta(f) = \max \limits_{\substack{P_1,P_2\in\mathcal{P}_{\mathcal{X}}\\ P_1\sim P_2}}{\left\lvert f(P_1)-f(P_2)\right\rvert}.
\]
\end{definition}

A standard way to obtain \ac{DP} for a numeric query $f$ is to add noise scaled by the sensitivity of $f$.

\begin{corollary}[Gaussian noise \cite{DworkRoth14}]\label{cor:GaussNoise}
For any $\varepsilon, \delta \geq 0$, %it suffices to 
sampling Gaussian noise $\eta \sim \mathcal{N}\left( 0, \frac{ 2 \ln \frac{1.25}{\delta} (\Delta(f))^2} {\varepsilon^2} \right)$
and adding this to $f(P)$ provides $(\varepsilon, \delta)$-\ac{DP} for $f$.
\end{corollary}

%\textbf{Please check: the variance should depend on $\Delta(f)^2  / \varepsilon^2$}
More generally, for any $\mathcal{P}'\subseteq \mathcal{P}_{\mathcal{X}}$, if $f \colon \mathcal{P}'\mapsto \mathbb{R}$ is a query on a subspace of $\mathcal{P}_{\mathcal{X}}$, the \emph{sensitivity of $f$ restricted to the subspace $\mathcal{P}'$} is defined as: 
\[
\Delta(f | \mathcal{P}') = \max_{\substack{P_{1}, P_{2} \in \mathcal{P}' \\ P_{1} \sim P_{2}}} {\left \lvert f(P_{1}) - f(P_{2}) \right \rvert}.
\]

\subsection{Federated Computation}\label{sec:fed_compute}
The federated model of computation has recently emerged to capture cases where multiple clients holding sensitive data, under the guidance of a coordinating server, collaboratively perform a computation over their data~\cite{Bharadwaj22}.
This covers a wide range of scenarios, from those where there are only a few powerful clients (\eg, hospitals), to where there are millions or billions of clients (\eg, mobile devices).
We will focus on the cases where there is a moderate to large number of relatively weak clients, representing mobile devices like phones or wearables, that are constrained in terms of computation or memory.
Within recent research advancements in federated computation, significant attention has focused on \ac{FL}, which aims to build a \ac{ML} model using clients' data.
\ac{FL} can be broadly classified into two categories: \textit{(i)} \emph{horizontal} or \emph{cross-device} \ac{FL}, where data is partitioned across clients, and all data points held by the clients share the same set of features (\eg, photos stored locally on clients' mobile phones), and \textit{(ii)} \emph{vertical} or \emph{cross-silo} \ac{FL}, where data is partitioned by features, making it relevant when different entities hold different attributes for the same set of users (\eg, hospitals in the same region specializing in different aspects of healthcare, each holding different pieces of health data for overlapping sets of patients).
In typical \ac{FL} use cases, where clients' data contain sensitive information (\eg, locations, health records, photos, text messages, etc.), it is natural to consider the associated data distribution of the clients as private.

\begin{definition}[Federated Model]
\label{def:trustmodel}
In this paper, we consider a horizontally federated environment consisting of a set of clients $C = \{c_{1},\ldots, c_{n}\}$ for some $n \in \mathbb{N}$ such that all clients independently and honestly comply with the protocol (\ie, no Byzantine clients nor any adversarial collusion between them).
The clients interact with a co-ordinating server $S$ to compute a function of interest.
The goal is to enable computation of functions of interest while minimizing the amount of sensitive information that is visible to the server.
\end{definition}
In order to achieve a good tradeoff between accuracy and trust, we adopt a ``trusted aggregator'' paradigm. 
%to circumvent the severe utility degradation inherent in strictly local differential privacy (LDP) regimes. Under a pure LDP setup, every participating client must independently perturb their local statistics before transmission. When estimating sensitive, non-linear global metrics like the Kullback-Leibler (KL) divergence, this localized noise accumulates additively across all users, overwhelmingly masking the true underlying data drift and rendering the resulting estimator practically useless. To achieve a viable utility-privacy trade-off, we offload the noise calibration to a trusted aggregation layer. 
Our protocols involve an aggregator which receives securely combined sampled client frequencies, computes the exact global estimate, and only then applies the requisite DP noise. 
That is, the aggregator obtains an already aggregated view of the client data, and merely completes the last step to turn it into the required estimator.  
Moreover, adopting a trusted aggregator does not strictly necessitate a centralized single server. 
In practice, this role can be decentralized and instantiated dynamically; for example, by randomly electing a trusted user node in each communication round, or by utilizing Secure Multiparty Computation (SMC) and Trusted Execution Environments (TEEs) to act as a virtual aggregator that guarantees code integrity and secure privacy noise injection.

We denote by \texttt{SecAgg} a secure aggregation, where the sum of data from a set of mutually distrustful sources is computed without revealing the value of any individual inputs.
Following Definition~\ref{def:trustmodel}, all clients are assumed to follow the protocol honestly, while the server is honest-but-curious.
Implementations of \texttt{SecAgg} range from a cluster-based model~\cite{Kim23} to a privacy-preserving protocol~\cite{Bonawitz17} that uses Shamir's $t$-out-of-$n$ secret sharing scheme~\cite{Shamir79} to account for dropout clients.
In \Cref{sec:PRIEST}, \texttt{SecAgg} will be used to securely aggregate the local datasets of the clients involved under different trust models.
This ensures that even the trusted aggregator only gets to see the combined information from all clients, rather than the raw data of any individual client. 

\paragraph{Threat model.}
Concretely, our threat model is as follows: 
clients hold their own data, never revealing it directly, and participating honestly in the protocol; 
a \texttt{SecAgg} protocol is used to collect statistics from subsets of clients, and reveals only the summation of its inputs; 
a trusted aggregator sees only the output of \texttt{SecAgg} and performs some small mathematical manipulation to produce a noisy version; 
the server sees only this noised output, and produces the final estimator for downstream use. 
All participants are honest-but-curious. 

\subsection{Monte Carlo subsampling}\label{sec:MC}
As probability distributions $P_{1}$ and $P_{2}$ are defined via observations of data, we are unable to compute the exact expression of their KL divergence as in \Cref{def:KLDiv}.
Instead, we will use Monte Carlo subsampling~\cite{Shapiro03}: 
build an estimator that takes in subsamples 
$x_{1}, x_{2}, \ldots \sim P_{2}$ and produces an estimator of $D_{\operatorname{KL}} \left(P_{1} \| P_{2} \right)$.
%A common method for constructing an accurate estimate of a sum is to use Monte Carlo subsampling~\cite{Shapiro03}.
%Given subsamples $x_{1}, x_{2}, \ldots \sim P_{1}$, is there an expression that each subsample $x_{i}$ can be substituted into, such that their sum estimates the KL divergence of $P_{1}$ and $P_{2}$?
Since KL divergence is a Bregman divergence~\cite{Amari09}, it can be measured as the shortest distance between a convex function and its tangent.
Setting $r_{i} = P_{1}(x_{i}) / P_{2}(x_i)$ for $x_i\sim P_{2}$, %of $P_{2}$, 
the estimator $\Phi(r_i) = (r_{i} - 1) - \ln{r_{i}}$
%\begin{equation}
%\Phi(r_i) = (r_{i} - 1) - \ln{r_{i}}
%\label{eq:estimator}
%\end{equation} 
measures the vertical distance between $\ln{r_{i}}$ and its tangent.
$\Phi(r_i)$ is the core of the estimator of KL divergence built in this paper. %in \Cref{alg:sys_model1,alg:sys_model2,alg:sys_model3} in \Cref{sec:PRIEST}.

\begin{figure*}[t]
\centering
\includegraphics[width=0.7\textwidth]{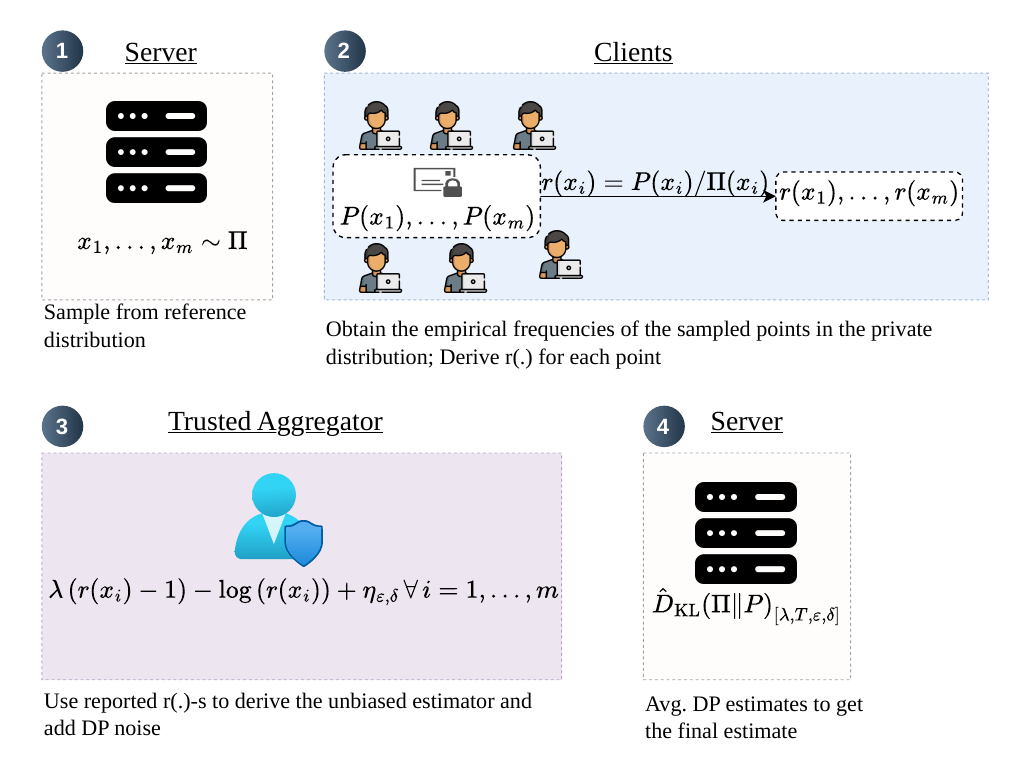}
\caption{The workflow of \sys with a trusted aggregator
}
\label{fig:workflow}
\end{figure*}

\section{DP estimator of KL divergence}\label{sec:PRIEST}

%\subsection{Problem definition and notations}

Given a collection of $n$ federated clients holding data from the space $\mathcal{X}$ which collectively defines a data set $D$, we wish to compute an accurate estimate of the KL divergence between this data distribution and a reference distribution with formal \ac{DP} guarantees.
That is, each client $i$ holds a local dataset $D_{i}$, so that $D(x) = \sum_{i=1}^{n} D_{i}(x)$.
The global dataset $D$ then defines a probability distribution $P$.
The server holds a public reference distribution $\Pi$, also over $\mathcal{X}$.
For instance, $\Pi$ could be derived from data collected in the past or in a controlled fashion using voluntary participation of users.
Hence, the goal is to   estimate $D_{\operatorname{KL}} \left( \Pi \| P \right)$ under privacy guarantees.

\paragraph{\sys in a nutshell.}
Our proposed approach, 
\sys, operates in three primary phases designed to securely and efficiently estimate distribution shifts. First, participating clients utilize a shared, public reference distribution ($\Pi$) to independently calculate the empirical frequencies of their local, private datasets over the reference support. Second, these localized statistics are securely transmitted to the trusted aggregator, which sums them to form a global empirical distribution representing the entire federation. 
Finally, the aggregator constructs a per-batch unbiased estimate of the KL divergence between this aggregated private distribution and the public reference. 
Utilizing rigorously analyzed and bounded sensitivity metrics, the aggregator then calculates and injects the necessary amount of calibrated differential privacy noise to privatize the estimate. 
This resulting privatized estimate of KL divergence is then shared back to clients or published, providing an accurate, privacy-preserving measure of systemic data drift without ever exposing any  user's data distribution. 
\Cref{fig:workflow} illustrates the high-level workflow of \sys with a trusted aggregator.
%To the best of our knowledge, this is the first approach, with a sound theoretical underpinning, to develop a privacy-preserving and unbiased estimate of KL divergence. 

To build-up to the design of \sys, we prove the following facts  %under the different trust models, we state and prove some useful facts 
that are needed to obtain the estimator of the KL divergence. 

\begin{fact}
\label{lem:estimator_fraction_expect}
If $\Pi \in \mathcal{P}_{\mathcal{X}}$ is a full-support distribution (\ie it excludes events with zero probability), for any random variable $X \sim \Pi$ and $P \in \mathcal{P}_{\mathcal{X}}$, we have $\mathbb{E} \left[r(X) \right] = 1$, where $r(X)=\frac{P(X)}{\Pi(X)}$.
\end{fact}

\begin{proof}
%Letting $r(X) = \frac{P(X)}{\Pi(X)}$: 
$\mathbb{E}[r(X)]$ can be expanded as
\begin{equation*}
\textstyle
    \sum\limits_{x \in \mathcal{X}} r(x) \Pi(x)= \sum\limits_{x \in \mathcal{X}} \frac{P(x)}{\Pi(x)}
    \Pi(x) = \sum \limits_{x \in \mathcal{X}} P(x) = 1. 
\end{equation*}    
\end{proof}

\begin{fact}
\label{lem:estimator_sum_expect}
For a random variable $X$ and a function $r$ on $X$, let $\Phi_{\Pi,P}(r(X),\lambda)=\lambda \left( r(X) - 1 \right) - \ln \left( r(X) \right)$. If $r(X)$ is as defined in \Cref{lem:estimator_fraction_expect} w.r.t. $\Pi$ and $P$, then, under the assumption of \Cref{lem:estimator_fraction_expect}, for any $\lambda \in \mathbb{R}$, we have $\mathbb{E}_\Pi \left[\Phi_{\Pi,P}(r(X),\lambda)\right] = D_{\operatorname{KL}} \left( \Pi \| P \right)$.
\end{fact}

\begin{proof}
%Letting $r(X) = \frac{P(X)}{\Pi(X)}$: 
As \Cref{lem:estimator_fraction_expect} implies $\mathbb{E}[r(X) - 1] = 0$, we get
\begin{align*}
&\mathbb{E}_\Pi \left[\Phi_{\Pi,P}(r(X),\lambda)\right]=\mathbb{E}_\Pi\left[ \lambda \left(r(X) - 1 \right) - \ln \left(r(X) \right) \right]\\ \nonumber 
&= - \mathbb{E}_\Pi[\ln{r(X)}]
= - \mathbb{E}_\Pi \left[ \ln{ \frac{P(x)}{\Pi(x)}} \right]= 
    %\mathbb{E} \left[ \ln{ \frac{\Pi(x)}{P(x)}} \right] = 
    D_{\operatorname{KL}}(\Pi \| P). \end{align*}
\end{proof}

%\begin{corollary}\label{corr:unbiased_est}
%\sayan{might be changed to a remark as this is a general property of unbiased estimators}
%    For $\lambda,T>0$, $r(X)$ as defined in \Cref{lem:estimator_fraction_expect}, and i.i.d. $X_1,\ldots,X_T\sim \Pi$, $\frac{1}{T}\sum_{t=1}^{T}\lambda(r(X_t)-1)-\ln\left(r(X_t)\right)$ is an unbiased estimator of $D_{\operatorname{KL}}(\Pi \| P)$.
%\end{corollary}

\begin{remark}\label{corr:unbiased_est}
    As \Cref{lem:estimator_sum_expect} establishes that
    %, for $X\sim \Pi$, 
    $\Phi(r(X),\lambda)$
    %, with $r(X)$ defined w.r.t. $\Pi$ and $P$ as before, 
    is an unbiased estimator of $D_{\operatorname{KL}}(\Pi\|P)$, it follows that for any arbitrary independent samples of $X_1,\ldots,X_M$ for some $M\in \mathbb{N}$, $\frac{1}{M}\sum_{i=1}^M \Phi(r(X_i),\lambda)$ is also an unbiased estimator of $D_{\operatorname{KL}}(\Pi\|P)$ with a factor $M$ decrease in its variance.  
\end{remark}

%\begin{proof}
%\begin{align*} 
%    &= \mathbb{E} \left[ \frac{1}{T} (\sum_{t = 1}^{T} \lambda(r(X_{t}) - 1) - \ln{r(X_{t})}) \right] \nonumber \\
%    &= \frac{T}{T} \mathbb{E}_{X \sim \Pi} \left[ \lambda_{0}(r(X) - 1) - \ln{r(X)}) \right]\nonumber\\
%    &\left[\text{as } x_{1}, \ldots, x_{T} \overset{\operatorname{i.i.d.}}{\sim} \Pi \right] \nonumber \\
%    &= D_{\operatorname{KL}(\Pi \|P)}. \nonumber \qquad(\text{using Facts } \ref{lem:estimator_fraction_expect} \text{ and } \ref{lem:estimator_sum_expect})
%\end{align*}\end{proof}

\Cref{corr:unbiased_est} enables the estimation of the KL divergence between $\Pi$ and $P$ by asking a subset of the clients to share the probability masses of a few points sampled by the server according to the reference distribution. %We now investigate the different settings in which the clients may be able to perturb the information they share with the server in this process to have a formal \ac{DP} guarantee on their locally held data while allowing the server to efficiently estimate the required KL divergence as correctly as possible. 
To make our estimation of KL divergence both communication-efficient and privacy-preserving, we propose a sampling-based approach that leverages this.
%the result derived in \Cref{corr:unbiased_est}. %We first present a fully distributed model of \sys with a straightforward way of directly adapting the derived estimator with local noise added by the clients to obtain the local \ac{DP} guarantee against an untrusted server, show its shortcomings, show that how having a trusted server would address these concerns, and extend our approach to propose an intermediate framework of \sys that upholds the statistical benefits of a centralized framework without needing to have a fully trusted server.  
%
%In order to have formal \ac{DP} guarantees, the information that the client shares with the ser locally held data points with the server.
For brevity, detailed proofs of all subsequent results concerning the theoretical analysis of \sys are deferred to the appendices. %extended technical report of this paper available at \texttt{https://arxiv.org/abs/2411.16478}.

\subsection{Baseline: pre-aggregation perturbation}\label{sec:sys_distributed}
We first present a canonical baseline that places no trust in the aggregator, by directly adapting the estimator derived in \Cref{corr:unbiased_est}.
In order to have \ac{DP} guarantees, the clients add noise to the aggregation of their individual records before the secure aggregation takes place to  
%\sayan{possibly cite something like DPrio here} 
share the sum with the server. 
%This straightforward approach proceeds as follows. 
In parallel, $T$ subsets of clients $C_t$ are selected to participate in each batch $t$ and are provided with i.i.d. points $x_{1_t},\ldots,x_{m_t}\in\mathcal{X}$ sampled by the server from the reference distribution $\Pi$. 
The clients then securely aggregate the frequencies of the $x_i$ computed over their local datasets and normalize them to obtain the empirical probability masses of $x_{1_t},\ldots,x_{m_t}$ in the subpopulation distribution, \ie $\Pr_{c \in C_t}\left[X = x_{1_t}\right],\ldots, \Pr_{c \in C_t}\left[X = x_{m_t}\right]$. 
The clients add independent Gaussian noise $\eta_{\varepsilon,\delta}$ to each reported value. 
The scale of the noise is determined by $\varepsilon$, $\delta$, and the number of items in $C_t$. If the same point is sampled more than once in a batch, the clients report its noisy frequency only once, and the server uses that value for every occurrence.
For a participating client $c$, let $D_c(x)$
denote the number of records at $c$ equal to $x$, and let $N_t = \sum_{c \in C_t} |D_c|$ denote the total number of records held by the clients participating in batch $t$. 
The empirical probability mass of $x$ in that batch is therefore $P_t(x) = 
\sum_{c\in C_t} D_c(x) / N_t$. 
%Specifically, writing $D_{c,d}$ for the number of times digit $d$ is seen by client $c$, let
%$N_{t} = \sum_{c \in C_t} D_{c,t}$. 
The distributed noise added by each client, summed over all clients, has sufficient variance to provide the desired $(\varepsilon,\delta)$-\ac{DP} guarantee. 
%is added to it using a distributed \ac{DP} noise adding mechanism such as DPrio~\cite{Keeler23} which provides 
% under its local model~\cite{DworkRoth14}.
The server receives the aggregated value,  $P$, and applies clipping to handle very small or negative values, obtaining
%Hence, a noisy version of the population probability masses of $x_1,\ldots,x_m$ in the population distribution, 
$P'(x_{i_t}) = \max\{ P(X = x_{i_t}) + \eta_{\varepsilon,\delta}, \, \tau \}$,
where $\tau > 0$ is a small constant. %introduced to ensure $P'(x_t) > 0$, is shared with the server. %\ie $P(X = x_t)$. In the fully distributed (\distributed) setting, where the server is untrusted, clients are required to ensure privacy locally by adding noise to any statistics they release. Accordingly, each client adds Gaussian noise $\eta_{\varepsilon,\delta}$, providing a local \ac{DP} guarantee~~\cite{DworkRoth14}, and shares
%$P'(x_t) = \max\{ P(X = x_t) + \eta_{\varepsilon,\delta}, \, \tau \}$,
%where $\tau > 0$ is a small constant introduced to ensure $P'(x_t) > 0$. 
This ensures that the logarithm computed by the server is well-defined, which is necessary for applying the KL divergence estimator from \Cref{corr:unbiased_est}.
%The full algorithm is presented in \Cref{alg:sys_dist}. 
The privacy properties of this baseline, presented in \Cref{alg:sys_dist}, follow by considering the sensitivity of the aggregated data reported, and scaling the Gaussian noise accordingly.
%The proof of this claim is deferred to the Appendix.

%The noise each client must add to their data to ensure \ac{DP} can be computed via its local model~\cite{DworkRoth14}.

%In this approach, the noise is added by the clients to guarantee \ac{DP}.

%The essence of our approach is to build an estimator for \kld by sampling a set of clients to participate in each round. The server samples 
%Each of these clients $C_{t}$ reports their frequencies, before the data from each round is securely aggregated and then sent to the server $S$.
%The $T$ rounds are completed in parallel to reduce the variance.
%The estimator, a convex continuous function of the ratio $r$ between the probability distributions $P$ and $\Pi$, is formed by $S$.
%We finally add $(\varepsilon, \delta)$-DP noise to the aggregate, to quantify how much information is disclosed about individual user records in the output.
%The full algorithm is presented in \Cref{alg:sys_model1}.

\begin{algorithm2e}[t]
    \DontPrintSemicolon
    \caption{\sys: Baseline}
    \label{alg:sys_dist}
    %\tcp{Client side:}
    \For{\text{batch }$t = 1, \dots, T$}{
    Sample (disjoint) set of participating clients $C_t$\;
    $x_{1_t},\ldots,x_{m_t} \overset{\operatorname{i.i.d.}}{\sim} \Pi$ and shared with $C_t$\; %clients in $C_t$\;
    \For{$i=1,\ldots,m$}{
    \For{$c\in C_{t}$}{
    \tcp{Client side:}
            $c$ finds frequency of $D_{c}(x_{i_t})$ \;
            %$P(x_{i_t})= \texttt{SecAgg}\left( \frac{\sum_{c \in C_{t}} D_{c}(x_{i_t})}{ N_{t}|}
            %+\eta_{\varepsilon, \delta} \right)$, where $\eta_{\varepsilon,\delta}\sim \mathcal{N}\left(\frac{2\ln{(1.25/\delta)}}{(N_{t}|\varepsilon)^2}\right)$ \;
            $P(x_{i_t})=\texttt{SecAgg}\left(\frac{\sum_{c\in C_t} D_c(x_{i_t})}{N_t} + \eta_{\varepsilon,\delta}\right),$
$\text{where } \eta_{\varepsilon,\delta} \sim \mathcal{N}\!\left(0,\ \frac{4\ln(1.25/\delta)}{(N_t\,\varepsilon)^2}\right)$\;
        }
    \tcp{Server side:}
            $P'(x_{i_t})=\max\{ P(x_{i_t}) ,\tau\}$ where
            %$N$ is no. samples $x_{i}$ in $D$, 
            $\tau > 0$ is small constant\;
            %$\eta_{\varepsilon, \delta}$ is $\left( \frac{\sqrt{2}\varepsilon}{N}, \delta \right)$-\ac{DP} noise, $N$ is no. samples $x_{i}$ in $D$, $\tau > 0$ is small constant \;
            $r'(x_{i_t}) = \frac{P'(x_{i_t})}{\Pi(x_{i_t})}$\;
            %Shared with the server 
            }
            }
    %$r'(x_{i}) = \frac{P'(x_{i})}{\Pi(x_{i})}$ $\forall\, i=1,\ldots,m_T$\;
    $\hat{D}_{\operatorname{KL}} \left( \Pi \| P \right)_{[\lambda, T,\varepsilon,\delta]} = \frac{1}{mT}\sum\limits_{t=1}^{T}\sum\limits_{i = 1}^{m}\Phi_{\Pi,P}(r'(x_{i_t}),\lambda)$
    %(\lambda(r'(x_{i}) - 1) - \ln{r'(x_{i}))}$\;
    %$D_{\operatorname{KL}} \left( \Pi \| P \right)^{\text{final}}_{\text{est}[\lambda_{0}, T, \varepsilon, \delta]} = D_{\operatorname{KL}} \left( \Pi \| P \right)^{\text{prelim}}_{\text{est}[\lambda_{0}, T]}$
\end{algorithm2e}

%\noindent
%\textbf{Analysis of the algorithm.}

\begin{theorem}
\label{th:Model3_DP}
The baseline version of \sys (\Cref{alg:sys_dist}) satisfies $(\varepsilon, \delta)$-\ac{DP}. 
\end{theorem}

%\graham{need to define example-level DP}

%However, we note that this straightforward approach where the clients are responsible for adding local noise and sending an locally obfuscated version of the aggregated probability masses of the points sampled from the reference distribution, although provides a baseline to evaluate the performance of \sys under \ac{DP} guarantees derived in the most canonical way, does not give an unbiased estimator of $D_{\operatorname{KL}}\left(\Pi\|P\right)$, due to the correction factor of $\tau$ to avoid negative values of the estimated ratio $r'(x_t)$.
%Therefore, a proof such as \Cref{corr:unbiased_est} would not be applicable, in turn, making the estimator undesirable from the perspective of utility, as the bias introduced by $\tau$ cannot be corrected by the server $S$ without compromising the \ac{DP} guarantees on the clients' inputs. 
%Hence, we proceed to propose a variant that allows for obtaining an unbiased estimator of the KL divergence using \Cref{corr:unbiased_est} and will treat this version of \sys, as presented in \Cref{alg:sys_dist}, as an empirical baseline against our improved estimator in the experimental analyses later on in this paper.

This straightforward approach, where clients add local noise and send a locally perturbed version of the aggregated probability masses of points sampled from the reference distribution, is useful as a baseline for evaluating the performance of \sys under canonical \ac{DP} guarantees.
However, it does not yield an unbiased estimator of $D_{\operatorname{KL}}\left(\Pi\|P\right)$. The bias arises from the correction factor $\tau$, introduced to avoid negative values in the estimated ratio $r'(x_{i_t})$.
Consequently, the error reduction offered via  \Cref{corr:unbiased_est} is not applicable.
This renders the estimator undesirable from a utility perspective, since the bias introduced by $\tau$ cannot be corrected by the server without violating the clients’ \ac{DP} guarantees. \Cref{alg:sys_dist} 
%version of \sys, presented in \Cref{alg:sys_dist}, 
will serve as an empirical baseline for comparison against our improved estimator. %in the experimental analyses later in this paper. 

\subsection{\sys with trusted aggregator}

%In light of the shortcomings of the baseline approach described in \Cref{sec:sys_distributed}, we now present the main version of \sys. The overall structure of the estimation algorithm remains similar; however, the tasks of frequency aggregation and noise injection (to ensure \ac{DP} guarantees) are now shared between the clients and a trusted aggregator. This contrasts with the earlier design, where all computations were handled locally by the clients, leading to bias in the final estimator. In this approach, we induce a slightly higher level of trust in the system and,  hence, introduce a trusted aggregator to carry out some of the privacy-aware operations before sharing the final terms to the server to carry out the final round of aggregation and derive the final estimate of $D_{\operatorname{KL}}(\Pi\| P)$. We remain consistent of our implicit threat model and objective of keeping the clients' raw data and the clear statistics performed on them private from the server.

In light of the shortcomings of the baseline approach described in \Cref{sec:sys_distributed}, we now present the main version of \sys in \Cref{alg:sys_tagg}. The overall structure of the estimation algorithm remains similar; however, the tasks of frequency aggregation and noise injection (required to ensure \ac{DP} guarantees) are now distributed between the clients and a trusted aggregator. 
%This design contrasts with the earlier approach, in which all computations were performed locally by the clients, leading to bias in the final estimator.
\sys makes a careful trust trade-off where securely aggregated sampled frequencies are processed by the trusted aggregator before being released to the server. 
This permits the non-linear transformation and calibrated DP noise addition to occur after secure aggregation, avoiding the bias introduced by perturbing the individual probability masses before the transformation.
%By leveraging a trusted aggregator, we make a slightly higher trust assumption but enable certain privacy-preserving operations to be executed more effectively before transmitting the resulting terms to the server. 
The server then performs the final aggregation step and derives the estimate of $D_{\operatorname{KL}}(\Pi \| P)$. 
This trust assumption can be discharged by performing the aggregation on secure hardware, such as via a Trusted Execution Environment (TEE)~\cite{Sabt_TEE_2015,Shepherd:Markantonakis:24}. 
Alternatively, it can be implemented in the secure multiparty computation model (SMC)~\cite{Cramer:Damgard:Nielsen:15}, where the main technical hurdle is to approximately compute the logarithm function~\cite{AlyS19}. %Throughout, 
%This remains consistent with our implicit threat model and the overarching objective of preventing the server from accessing clients’ raw data or the clear statistics derived from them as outlined in \Cref{sec:fed_compute}. A more detailed discussion on the role of the trusted aggregator and the trust conferred upon it has been given by \Cref{rem:trusted_agg} at the end of \Cref{sec:PRIEST}.
This is consistent with the threat model of \Cref{sec:fed_compute}, whose central objective is to keep clients' raw data and any statistics derived from it in the clear hidden from the server. 
This narrow trust assumption is practical
and minimizes the trusted work needed:  the aggregator only evaluates a single logarithm and performs basic arithmetic on securely aggregated scalars.
This point is discussed further in \Cref{rem:trusted_agg} at the end of \Cref{sec:PRIEST}.

%\textbf{Overview of the algorithm.}
%Our approach is similar to the \fulltrust model, in that noise is still added to the estimator to guarantee DP.

\begin{algorithm2e}[t]
    \DontPrintSemicolon
    \caption{\sys with trusted aggregator}
    \label{alg:sys_tagg}
   \For{\text{batch }$t = 1, \dots, T$}{
    Sample (disjoint) set of participating clients $C_t$\;
    $x_{1_t},\ldots,x_{m_t} \overset{\operatorname{i.i.d.}}{\sim} \Pi$ and shared with $C_t$\; 
    \For{$c \in C_t$}{
    \tcp{Client side:}
        \For{$i=1,\ldots,m$}{
            $c$ finds frequency of $D_{c}(x_{i_t})$\;
            $P(x_{i_t}) = \texttt{SecAgg} \left( \frac{ \sum_{c \in C_{t}} D_{c}\left(x_{i_t}\right)}{N_{t}} \right)$\;
            $r(x_{i_t}) = \frac{P(x_{i_t})}{\Pi(x_{i_t})}$\;
            Shared with the trusted aggregator\;
        }
    }
    }
    \tcp{Trusted aggregator side:}
    %$r(x_{i}) = \frac{P(x_{i})}{\Pi(x_{i})}$ $\forall i=1,\ldots,$\; 
    \For{$t=1,\ldots,T$}{
        \For{$i=1,\ldots,m$}{
        %$\Phi'_{\Pi,P}(r(x_{i_t}),\lambda)=\Phi_{\Pi,P}(r(x_{i_t}),\lambda)+\eta_{\varepsilon,\delta}$, 
        %s.t. $\eta_{\varepsilon,\delta}\sim \mathcal{N}\left(0,\frac{2\ln(1.25/\delta)\Delta\left(\hat{D}_{\operatorname{KL}}(\Pi\| P)_{[\lambda,T]}\right)^2}{(\varepsilon T)^2}\right)$\;
        $\Phi'_{\Pi,P}(r(x_{i_t}),\lambda) = \Phi_{\Pi,P}(r(x_{i_t}),\lambda) + \eta_{\varepsilon,\delta}$, s.t.
$\eta_{\varepsilon,\delta} \sim \mathcal{N}\!\left(0,\ \frac{2\ln(1.25/\delta)\, m\, \kappa^2}{\varepsilon^2}\right)$,
$\kappa = \frac{|\lambda|}{N_t\pi_{\min}} + \ln 2$\;
        Shared with the server\;
        }
    }
    %$A = \sum_{t = 1}^{T} {\ln} \left( {r(x_{t})} \right) + \eta'_{\varepsilon, \delta}$ with $S$\;
    %$B = \sum_{t = 1}^{T} \lambda(r(x_{t}) - 1) + \eta'_{\varepsilon, \delta}$ with $S$\;
    %[where $\eta'_{\varepsilon, \delta}$ is $\left( \frac{\sqrt{2} \varepsilon}{T}, \delta \right)$-DP noise]\;
    \tcp{Server side:}
     $\hat{D}_{\operatorname{KL}} \left( \Pi \| P \right)_{[\lambda, T,\varepsilon,\delta]} = \frac{1}{mT}\sum\limits_{t=1}^{T}\sum\limits_{i_t = 1}^{m}\Phi'_{\Pi,P}(r(x_{i_t}),\lambda)$
\end{algorithm2e}

%\sayan{First add theorem on unbiased, bounded variance, then privacy. This way, we'll be able to give a bound on the sensitivity}

%\subsubsection{Analysis of the algorithm}
\label{sec:tagg_analysis}
%the two models discussed above, the analysis of the output and the corresponding sensitivity analysis (essential for satisfying DP) is identical across the two settings~\footnote{\aggregator essentially ``splits'' the noise in two different segments to foster the same DP guarantee as \fulltrust.}.

%Therefore, in this section, we carry out the privacy analysis catering to both \aggregator and \fulltrust.

\noindent{\textbf{Privacy analysis.}}
%\subsubsubsection{\textbf{Sensitivity analysis}}
To support the privacy analysis of our algorithm, we first bound the sensitivity of the estimator that is at the heart of \Cref{alg:sys_tagg}. 
%Although the exact heuristics of adding noise for privacy may differ across applications, it is crucial to obtain a bound on the sensitivity of our estimate in order to have the \ac{DP} guarantees expression derived in \Cref{alg:sys_tagg} well-defined. 
In what follows, let $\hat{D}_{\operatorname{KL}}(\Pi\|P)_{[\lambda,T,\varepsilon,\delta]}$ denote the estimate of the true value of $D_{\operatorname{KL}}(\Pi\|P)$ derived by \sys. 
In the end, the accuracy and privacy guarantees of the estimation are affected by the \emph{precision parameter} $T \in \mathbb{N}$, the \emph{variance parameter} $\lambda\in\mathbb{R}$, and the $\varepsilon$ and $\delta$ parameters for achieving $(\varepsilon, \delta)$-\ac{DP}.

%For any $P \in \mathcal{P}_{\mathcal{X}}$ representing the data distribution of participating clients and $\Pi \in \mathcal{P}_{\mathcal{X}}$ serving as the reference (non-private) distribution to be compared with $P$, 

For convenience, let $f(P) = D_{\operatorname{KL}}(\Pi \| P)$ %(\ie the true KL divergence between $\Pi$ and $P$) 
and $\hat{f}(P) = \hat{D}_{\operatorname{KL}} \left( \Pi \| P \right)_{[\lambda, T]}$, where $\hat{D}_{\operatorname{KL}}(\Pi\|P)_{[\lambda,T]}$ is the estimate of $D_{\operatorname{KL}}(\Pi\| P)$ without the \ac{DP} noise. % (\ie the estimated KL divergence between $\Pi$ and $P$ given by \Cref{alg:sys_tagg}).
We now analyze %the impact of the \ac{DP} noise on 
the sensitivity of $f$ and $\hat{f}$ as a precursor to appropriately add noise to obtain the final version of our estimator, $\hat{D}_{\operatorname{KL}}(\Pi\|P)_{[\lambda,T,\varepsilon,\delta]}$, with $(\varepsilon,\delta)$-\ac{DP} guarantee. Let $\mathcal{H}_{\mathcal{X}}^{>1}$ %$\mathcal{H}_{\mathcal{X}}^{>1} = \{h \in \mathcal{H}_{\mathcal{X}} \colon h(x) > 1 \forall x \in \mathcal{X} \}$ 
be the space of datasets where each value from $\mathcal{X}$ has frequency more than one.
Correspondingly, let $\mathcal{P}^{>1}$ be the set of probability distributions such that
$
\mathcal{P}^{>1} = \left\{ P \in \mathcal{P}_{\mathcal{X}} \colon \, \psi^{-1}(P) \in \mathcal{H}_{\mathcal{X}}^{>1} \right\}
$. 
This restriction is used in  the proof to avoid the vacuous case of an unbounded sensitivity value.
%handles a common issue in \ac{ML} settings, where events with zero or near-zero probability would create an unbounded error; restricting our attention to this case can be understood as implicitly applying a smoothing term. 
Accordingly, we derive a constant upper bound on the sensitivity of $f$ as follows.

\begin{restatable}{theorem}{SensitivityOfKLD}~\label{th:SensitivityOfKLD}
    For $\mathcal{H}^{>1}_{\mathcal{X}}$ with a fixed cardinality,
    $\Delta \left( f| \mathcal{P}^{>1} \right) < 0.7$. 
\end{restatable}

%The analysis proceeds by upper bounding the  change in KL divergence between neighboring distributions, by considering the worst-case values of the probability distribution. 
%The details, and those of subsequent proofs, are deferred to the Appendix for brevity. 
%The finite-sensitivity result assumes positive support with aggregate counts exceeding one on the effective estimation domain. 
%This is the standard support issue inherent in $\hat{D}_{\operatorname{KL}}(\Pi \| P)$: if $\Pi(x)>0$ while $P(x)=0$, the divergence itself is infinite. 
%In settings with rare categories, the protocol can therefore be applied to a prespecified smoothed/effective distribution, in which case the target is the KL divergence of that distribution. 
%\sys's estimation and privacy analysis then apply unchanged.
%While bounding the sensitivity to a constant requires restricting the domain to $\mathcal{P}^{>1}$ (where every element occurs more than once), this condition is easily met in practice. 
%Any singletons in the aggregated histogram can be eliminated via standard Laplace smoothing at a negligible cost to accuracy, without altering the core \sys protocol.
%Next, we bound the sensitivity of the estimated KL divergence with \ac{DP} guarantee, as given by \Cref{alg:sys_tagg}.

The analysis proceeds by upper bounding the change in KL divergence between neighboring
distributions, by considering the worst-case values of the probability distribution. The
finite-sensitivity result assumes positive support with aggregate counts exceeding one on the
effective estimation domain. This is the standard support issue inherent in
$D_{\mathrm{KL}}(\Pi\|P)$: if $\Pi(x)>0$ while $P(x)=0$, the divergence itself is not defined. \Cref{rem:min_occurance} shows how we can apply smoothing to lift this restriction. 
First, we bound the sensitivity of the
estimated KL divergence with DP guarantee, as given by \Cref{alg:sys_tagg}.

\begin{restatable}{theorem}{SensitivityOfSampledKLD}\label{th:SensitivityOfSampledKLD}
For $\mathcal{H}^{>1}_{\mathcal{X}}$ with a fixed cardinality $N$, we have %if $\lambda$ is the variance parameter used in the estimation and $\pi_{\min} = \min_{x \in \mathcal{X}} \Pi(x)$, we have $\textstyle
%\Delta \left( \hat{f} \hspace{0.05em} | \hspace{0.05em} \mathcal{P}^{>1} \right) \leq \frac{|\lambda|}{N\pi_{\min}} + \ln{2}$.
%\[
%\Delta \left( \hat{f} \hspace{0.05em} | \hspace{0.05em} \mathcal{P}^{>1} \right) \leq (\hat{\alpha} - 1) \hspace{0.05em} | \hspace{0.05em} \lambda_{P_{1}} - \lambda_{P_{2}}| \hspace{0.1em} + \hspace{0.1em} \frac{\lambda_{P_{2}}}{Np_{\min}} + \ln{2},
%\]
$\textstyle
\Delta \left( \hat{f} \hspace{0.05em} | \hspace{0.05em} \mathcal{P}^{>1} \right) \leq \frac{|\lambda|}{N\pi_{\min}} + \ln{2}$,
%\[
%\textstyle
%\Delta \left( \hat{f} \hspace{0.05em} | \hspace{0.05em} \mathcal{P}^{>1} \right) \leq \frac{|\lambda|}{N\pi_{\min}} + \ln{2},
%\]
where $\lambda$ is the variance parameter used in the estimation and $\pi_{\min} = \min_{x \in \mathcal{X}} \Pi(x)$.
%\[
%\hat{\alpha} = \max \left\{ \max_{x \in \mathcal{X}} \frac{P_{1}(x)}{\Pi\hspace{0.05em}(x)}, \hspace{0.1em} \max_{x \in \mathcal{X}} \frac{P_{2}(x)}{\Pi\hspace{0.05em}(x)} \right\}\text{ and }p_{\min} = \min_{x \in \mathcal{X}} \Pi\hspace{0.05em}(x).
%\]
\end{restatable}

The desired privacy guarantee then follows since the algorithm adds noise as a function of the computed sensitivity.
%; the proof is in supplementary material. 

\begin{restatable}{theorem}{ModelTwoDP}\label{th:Model2_DP}
The output of \sys with a trusted aggregator (\Cref{alg:sys_tagg}) satisfies  $(\varepsilon, \delta)$-\ac{DP}.
\end{restatable}

\begin{remark}\label{rem:min_occurance}
The assumption that $\mathcal{P}^{>1}$ is made to simplify the analysis of \sys, and can be easily lifted when applying a standard smoothing technique. 
In particular, for a constant $s>0$ fixed
independently of the data, the aggregator can instead use
$\tilde{P}_t(x)=\frac{N_tP_t(x)+s}{N_t+s|\mathcal{X}|}$
in place of $P_t(x)$ when forming the ratio $r$. 
Now any substitution between neighboring inputs moves smoothed counts of 
two values that are at least $s$ and differ by one, so every log-ratio changes by at most
$\ln(1+1/s)$ for all adjacent datasets.
This holds whatever the counts are.
The bound on the $\lambda$-term is unaffected. 
The bounds of \Cref{th:SensitivityOfKLD,th:SensitivityOfSampledKLD} thus become $\ln(1+1/s)$ and
$\frac{|\lambda|}{N_t\pi_{\min}}+\ln(1+1/s)$ without restricting the datasets, and the privacy guarantee of \sys (\Cref{th:Model2_DP}) holds
with $\kappa=\frac{|\lambda|}{N_t\pi_{\min}}+\ln(1+1/s)$ in the noise of \Cref{alg:sys_tagg}. 
Choosing $s=1$ corresponds to the standard Laplace smoothing.
Then $\ln(1+1/s)=\ln 2$ and the constants are exactly those obtained above for
$\mathcal{P}^{>1}$. 
For the results below on utility, 
replacing $\ln 2$ by $\ln(1+1/s)$, and $\ln 4$ by $2\ln(1+1/s)$ extends
\Cref{th:gen_estimator_variance} and \Cref{thm:opt-var} to any $s$. 
A smaller $s$ distorts the target less, as $\tilde{P}_t$ mixes $P_t$ with the uniform distribution with weight $\frac{s|\mathcal{X}|}{N_t+s|\mathcal{X}|}$, but
requires more noise.
\end{remark}

\noindent{\textbf{Utility analysis.}}
%We now present the formal analysis of various utility aspects of the proposed algorithm. In particular, 
Next, we establish that the estimate given by \Cref{alg:sys_tagg} is unbiased at the batch-level and bound the behaviour of its variance as a function of the parameter $\lambda$.
%associated with the algorithm. 

%\begin{restatable}{theorem}{estimatorunbiased}\label{th:estimator_unbiased}
%\sys with a trusted aggregator gives an unbiased estimator of $D_{\operatorname{KL}}(\Pi \| P)$.
%\end{restatable}

\begin{theorem}\label{th:estimator_unbiased}
Let $P_t$ denote the empirical distribution of batch $t$. \textsc{FedPriKL} with a trusted aggregator satisfies
\[
\mathbb{E}\big[\hat{D}_{\mathrm{KL}}(\Pi\|P)_{[\lambda,T,\varepsilon,\delta]}\big]
=\frac{1}{T}\sum_{t=1}^{T}D_{\mathrm{KL}}(\Pi\|P_t).
\]
%giving an unbiased estimator of the KL divergence.
\end{theorem}

This follows by applying  \Cref{lem:estimator_fraction_expect,lem:estimator_sum_expect} in each batch, where $r=P_t/\Pi$, since the noise has zero mean.
For equal-size batches, $P$ is the average of the $P_t$, so by convexity of $D_{\mathrm{KL}}$ in its second argument
the right-hand side is at least $D_{\mathrm{KL}}(\Pi\|P)$, with equality iff the batch distributions coincide.
%Hence the estimator is unbiased for $D_{\mathrm{KL}}(\Pi\|P)$ when $T=1$, and otherwise overestimates it in
%expectation by a gap that shrinks as the batches grow. The remaining results are stated for a common batch distribution $P_t=P$.
Hence the estimator is unbiased for $D_{\mathrm{KL}}(\Pi\|P)$ when $T=1$ or when the batch distributions coincide, and otherwise overestimates it in expectation. The remaining results are stated for equal-size batches of $N$ records with a common distribution $P_t=P$.

%In addition to the estimator being unbiased, we show that it also has a bounded variance.
%in the following theorem.
%(in the interest of space, the detailed proof is provided in the supplementary material). 

\begin{restatable}{theorem}{modeltwoestimatorvariance}\label{th:gen_estimator_variance}
    If $P$ and $\Pi$ are full-support distributions, let $N$ and $\pi_{\min}$ be as in \Cref{th:SensitivityOfSampledKLD}, and $\alpha=\max_{x\in\mathcal{X}}\frac{P(x)}{\Pi(x)}$ and $\beta=\max_{x\in\mathcal{X}}\frac{\Pi(x)}{P(x)}$. Then the estimator given by \sys in \Cref{alg:sys_tagg} % with a trusted aggregator 
    ensures:
    \begin{align*}&\operatorname{Var} \left[ \hat{D}_{\operatorname{KL}} \left( \Pi \| P \right)_{[\lambda, T, \varepsilon, \delta]} \right]\\ \nonumber
    &\leq \frac{1}{mT}\left(\lambda^{2} (\alpha -1) 
    + \max\{\alpha,\beta^2\} -1 -D_{\operatorname{KL}}(\Pi\|P)^2 \right.\\
    &\left.- 2\lambda\left(D_{\operatorname{KL}}(P \| \Pi)+ D_{\operatorname{KL}}(\Pi \| P)\right)\right)\\
   &+\frac{1}{T\varepsilon^2}\left(2\ln(1.25/\delta)\left((N\pi_{\min})^{-1}|\lambda|+\ln{2}\right)^2\right).
    \end{align*}
\end{restatable}

The subsequent section
%\Cref{sec:opt_lambda} in the appendix 
provides a detailed analysis of the optimal choice of the variance parameter $\lambda$ that minimizes the variance of the estimator produced by \sys.

\noindent{\textbf{Complexity analysis.}} 
%Finally, we dissect the computational complexity of \sys as presented in \Cref{alg:sys_tagg}. 
We sample all $n$ clients partitioned in $T$ disjoint batches. 
In each of the $T$ batches, $m$ instances of $X\sim\Pi$ are sampled, which the clients participating in the corresponding round operate with. 
The computational complexity associated with deriving and performing a secure-aggregation-based averaging on the frequencies for each of the $m$ sampled points is taken to be linear. 
%ignored as these methods could be orthogonally opmized w.r.t. \sys. 
Hence, the client-side computational complexity of \Cref{alg:sys_tagg} is $\mathcal{O}(nm)$. 
%trusted aggregator would be needed to 
On the trusted aggregator, the complexity of adding the Gaussian noise independently to each of the $mT$ points and taking the average is $\mathcal{O}(mT)$. 
%As the averaging takes server $\mathcal{O}(1)$ time
Thus, the overall algorithmic complexity of \sys in \Cref{alg:sys_tagg} is $\mathcal{O}(m(n+T))$. 

%\sayan{@Graham: please check! Do we also need to add the complexity of the full sharing, at least on the clients' side?}

%it seems to me that we are O(nm) [as we are sampling m instances of x~\Pi in each of the T batches, and we exhaust all n clients over the T batches sampling them disjointly]. If full-sharing was happening, it would have been O(n|X|), where |X|= total size of the input alphabet, and , of course m << |X|?

\begin{figure*}[t]
\centering
\subfloat[$\varepsilon = 0.1$]{
    \includegraphics[width=0.33\textwidth]{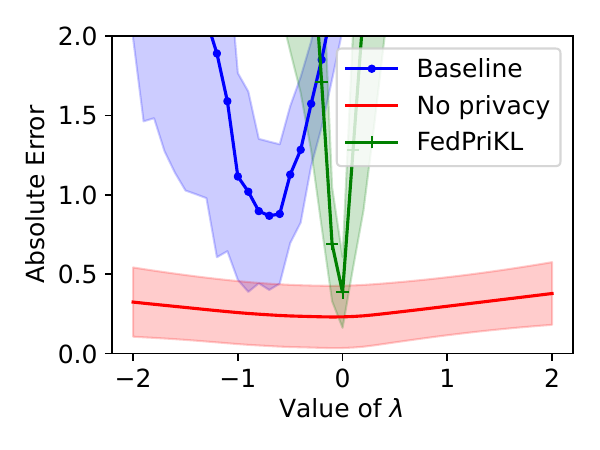}\label{fig:lambda0.1}}
\subfloat[$\varepsilon = 0.5$]{
    \includegraphics[width=0.33\textwidth]{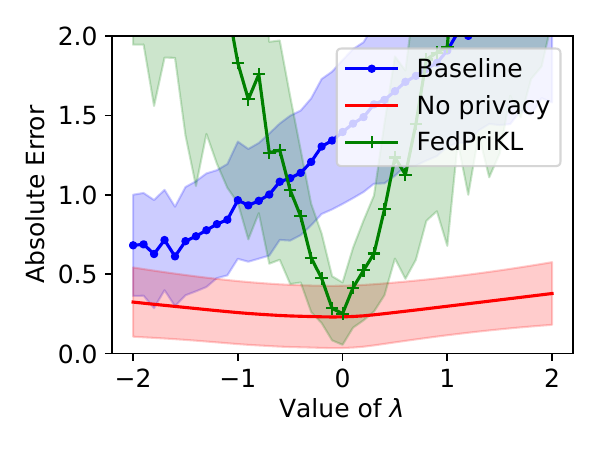}\label{fig:lambda0.5}}
\subfloat[$\varepsilon = 1.0$]{
    \includegraphics[width=0.33\textwidth]{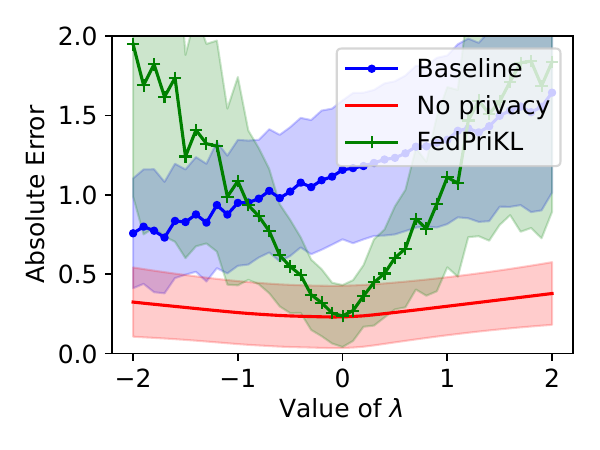}\label{fig:lambda1.0}
}
\caption{Absolute error of different estimators as $\lambda$ varies for $\epsilon = 0.1, 0.5, 1.0$}
\label{fig:lamdaeps}
\end{figure*}

\begin{figure}
\centering
%\hspace*{-2mm}
\subfloat[$|C_t|$ = 71 ($\sim 2\%$ of clients)]{
\includegraphics[width=0.45\columnwidth]{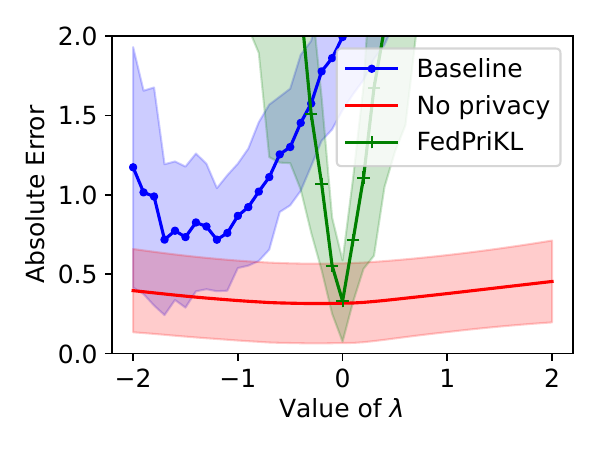}
\label{fig:Ct-1}
}
%\subfloat[$|C_t|$ = 179 ($\sim$ 5\% of clients)]{
%\includegraphics[width=0.33\textwidth]{figs/exp2-samples=179.pdf}
%\label{fig:Ct-5}
%}
\subfloat[$|C_t|= 358 (\sim10\%$ of clients)]{
\includegraphics[width=0.45\columnwidth]{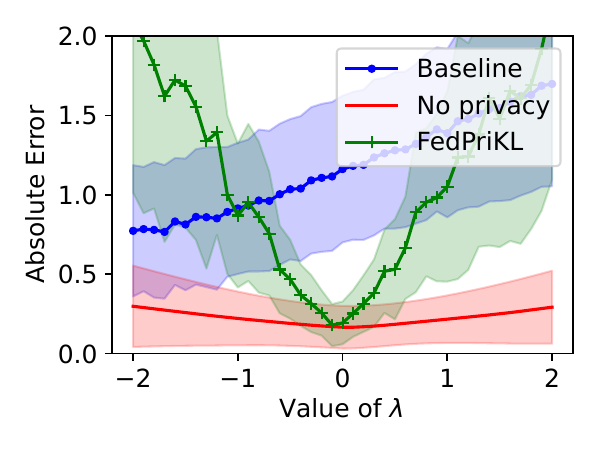}
\label{fig:Ct-10}
}
\caption{Absolute error for different batch sizes $|C_t|$. }
\label{fig:Ct}
\end{figure}

Complementing the discussion in \Cref{rem:min_occurance} on item frequency assumptions, we briefly discuss another core technical requirement for \sys that is crucial to its privacy and accuracy guarantees, namely the role of the trusted aggregator. 

\begin{remark}\label{rem:trusted_agg}
    We emphasize that the trust placed in the aggregator is narrow in
    both scope and exposure. Functionally, the aggregator performs exactly one
    step of \Cref{alg:sys_tagg} (line~14): applying $\Phi$ to the securely aggregated
    ratios and perturbing the result with Gaussian noise calibrated to the
    sensitivity bound of \Cref{th:SensitivityOfSampledKLD}. It never accesses raw client data and, aligned with the
    threat model of \Cref{sec:fed_compute}, \texttt{SecAgg} ensures that the aggregator's
    entire view is limited to the empirical masses of the $m$ points sampled from
    $\Pi$ in each batch, already summed over the participating clients of that
    batch, \ie $mT$ scalars in total, as opposed to individual datasets or full
    histograms over $\mathcal{X}$. Moreover, since the delegated task amounts to
    a few arithmetic operations and one logarithm evaluation, the aggregator need
    not exist as a distinguished physical party.  It can be instantiated inside a
    TEE~\cite{Sabt_TEE_2015,Shepherd:Markantonakis:24}; emulated via SMC~\cite{Cramer:Damgard:Nielsen:15}, for which protocols to securely
    evaluate the logarithm are well studied~\cite{AlyS19}; or realized under
    fully homomorphic encryption (FHE)~\cite{Gentry_FHE_2009} based techniques. 
    It is precisely the simplicity of the
    aggregator's task that makes such instantiations practical.
    For this reason, it is unnecessary to dwell on the implementation of the aggregator, since our focus is on the theoretical behavior of \sys.
%    optimizing it with advanced and secure machinery remains orthogonal to the core functionality of \sys.
\end{remark}

\begin{figure}[t]
\centering
\hspace{-3mm}\subfloat[$\varepsilon = 0.1$]{
\includegraphics[width=0.45\columnwidth]{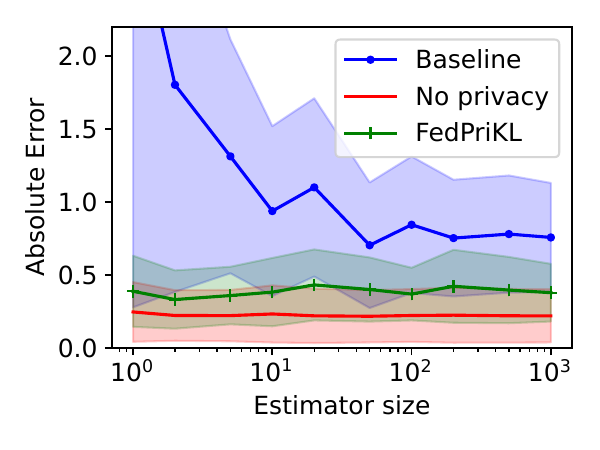}
\label{fig:meps0.1}
}
%\subfloat[$\varepsilon = 0.5$]{
%\includegraphics[width=0.33\textwidth]{figs/exp4-eps=0.5.pdf}
%\label{fig:meps0.5}
%}
\subfloat[$\varepsilon = 1.0$]{
\includegraphics[width=0.45\columnwidth]{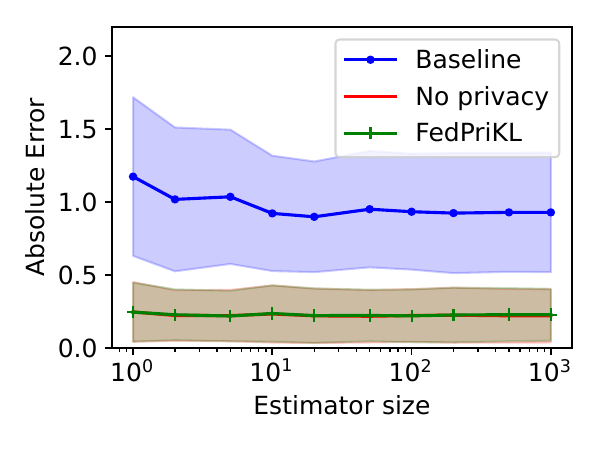}
\label{fig:meps1.0}
}
\caption{\mbox{Absolute error varying number of samples $m$}}
\label{fig:m}
\end{figure}

\begin{figure*}[t]
\centering
\subfloat[$KL(3,8)$ -- closest pair]{
\includegraphics[width=0.33\textwidth]{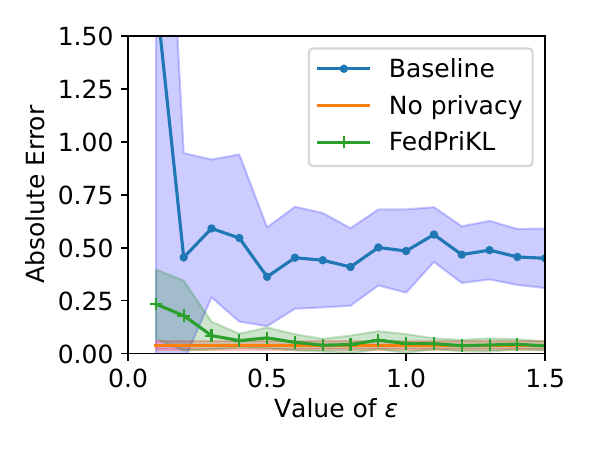}
\label{fig:pair38}
}
\subfloat[$KL(4,1)$ -- median pair]{
\includegraphics[width=0.33\textwidth]{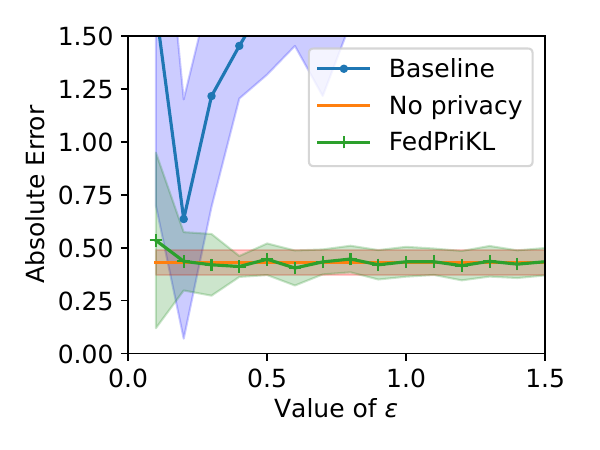}
\label{fig:pair41}
}
\subfloat[$KL(9,0)$ -- furthest pair]{
\includegraphics[width=0.33\textwidth]{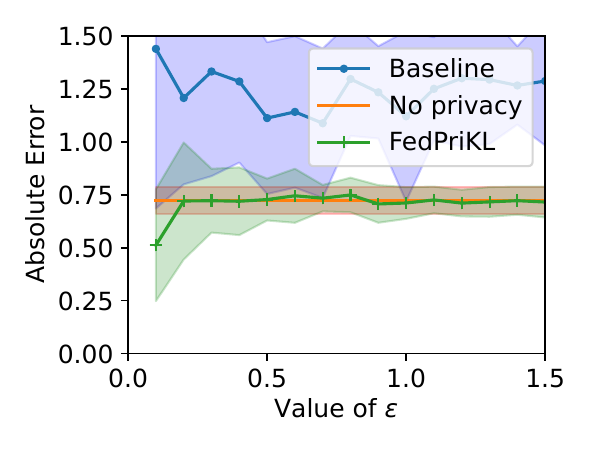}
\label{fig:pair90}
}
\caption{Absolute error of different estimators for different digit pairs}
\label{fig:pair}
\end{figure*}

%\sayan{on assuming every item occurs >1: if a point occurs 0 times we can discard, un;ikely that a point occurs exactly once across all clients. if it does, we can apply some smoothening techniques (e.g., Laplace correction). This problem remains orthogonal to our algorithm. sensitivity becomes unbounded in case where this happens, so we had to make this assumption. }

%\sayan{on having a trusted aggregator: limited task for the trusted entity, just one line of the algorithm. they're not getting all data, seeing someprobablities of the few items of the few clients aggregated across clients. This can also be done using some TEE, or FHE or MPC as the task assigned to the trusted aggregator is simple. }

\section{Analysis of the optimal variance parameter}\label{sec:opt_lambda}
%\sayan{@Graham: pleae review}

We note that the bounds of the variance of our estimator depend on the choice of the parameter $\lambda$.
Therefore, we seek a choice of $\lambda$ that minimizes the variance as follows.

\begin{definition}[Optimal variance parameter]\label{def:opt_var_param}
    We call a variance parameter $\lambda_{0, P}$ \emph{optimal} for $P$ under \sys with a trusted aggregator if, for every $\lambda \in \mathbb{R}$,
    \[
    \operatorname{Var} \left[ \hat{D}_{\operatorname{KL}} \left( \Pi \| P \right)_{[\lambda_{0, P}, T, \varepsilon, \delta]} \right] \leq \operatorname{Var} \left[ \hat{D}_{\operatorname{KL}} \left( \Pi \| P \right)_{[\lambda, T, \varepsilon, \delta]} \right].
    \]
\end{definition}

Now we present a theoretical analysis  to provide a foundational characterization and understanding on the optimal variance parameter $\lambda$ which, in turn, optimizes the estimate obtained with \sys. 
For notational convenience, we omit $P$ from the subscript of $\lambda_{0,P}$ and refer to the optimal variance parameter simply as $\lambda_{0}$, as the clients' data distribution $P$ is fixed in this context. 
In what follows, all expectations and variances are taken over $X\sim\Pi$, unless otherwise specified. 
Hence, the subscript $X\sim \Pi$ can be dropped from $\mathbb{E}_{X\sim \Pi}[.]$ and $\operatorname{Var}_{X\sim \Pi}[.]$ and we simply write $\mathbb{E}[.]$ and $\operatorname{Var}[.]$ when $X\sim\Pi$ is clear in the context.

\begin{theorem}\label{th:opt_lambda_no_noise}
    Under the assumptions of \Cref{th:gen_estimator_variance}, if \emph{no} \ac{DP} noise was introduced in \sys, we have $\lambda_0=\frac{D_{\operatorname{KL}}(P \| \Pi)+ D_{\operatorname{KL}}(\Pi \| P)}{\operatorname{Var}_{X\sim\Pi}[r(X)]}$, where $r(X)$ is as defined in \Cref{lem:estimator_fraction_expect}.
\end{theorem}

\begin{corollary}\label{corr:opt_lambda_noise}
    Under the same assumptions %of \Cref{th:opt_lambda_no_noise}, 
    we have
    %$$\lambda_0\geq \max\left\{0,\frac{D_{\operatorname{KL}}(P \| \Pi)+ D_{\operatorname{KL}}(\Pi \| P)}{\alpha -1}\right\}.$$
    $\lambda_0\geq \frac{D_{\operatorname{KL}}(P \| \Pi)+ D_{\operatorname{KL}}(\Pi \| P)}{\alpha -1}$, where $\alpha$ is as defined in \Cref{th:gen_estimator_variance}.
\end{corollary}

\begin{comment}
\begin{theorem}\label{th:opt_lambda_noise}
Under the assumptions of \Cref{th:gen_estimator_variance}, for \sys with a trusted aggregator, we have the following:
\begin{enumerate}
    \item If $D_{\operatorname{KL}}(P \| \Pi)+ D_{\operatorname{KL}}(\Pi \| P)\geq\frac{\ln{(1.25/\delta)\ln{4}}}{(T\varepsilon)^2N \pi_{\min}}$, the optimal $\lambda_0\in \mathbb{R}_{\geq 0}$ is s.t.
    $$\lambda_0=\frac{D_{\operatorname{KL}}(P \| \Pi)+ D_{\operatorname{KL}}(\Pi \| P)-\frac{\ln{(1.25/\delta)\ln{4}}}{(T\varepsilon)^2N \pi_{\min}}}{\operatorname{Var}[r(X)]+\frac{2\ln{(1.25/\delta)}}{(T\varepsilon N \pi_{\min})^2}}.$$

    \item If $\operatorname{Var}_{X\sim \Pi}[r(X)]< \frac{2\ln{(1.25/\delta)}}{(T\varepsilon N \pi_{\min})^2}$, the optimal $\lambda_0\in \mathbb{R}_{<0}$ is s.t.

    $$\lambda_0=\frac{D_{\operatorname{KL}}(P \| \Pi)+ D_{\operatorname{KL}}(\Pi \| P)+\frac{\ln{(1.25/\delta)\ln{4}}}{(T\varepsilon)^2N \pi_{\min}}}{\operatorname{Var}_{X\sim \Pi}[r(X)]-\frac{2\ln{(1.25/\delta)}}{(T\varepsilon N \pi_{\min})^2}}.$$

\end{enumerate}   
\end{theorem}
\end{comment}

\begin{theorem}\label{thm:opt-var}
Under the assumptions of \Cref{th:gen_estimator_variance}, for \sys with a trusted aggregator,
the optimal variance parameter is $\lambda_0=\max\{0,\lambda^\star\}$, where
$
\lambda^\star=\frac{D_{\mathrm{KL}}(P\|\Pi)+D_{\mathrm{KL}}(\Pi\|P)-\frac{m\ln(1.25/\delta)\ln 4}{\varepsilon^2N\pi_{\min}}}
{\mathrm{Var}[r(X)]+\frac{2m\ln(1.25/\delta)}{(\varepsilon N\pi_{\min})^2}}
$. In particular, $\lambda_0=0$ whenever the numerator is non-positive.
\end{theorem}

\begin{corollary}\label{corr:opt_lambda_order}
    $|\lambda_0|$ for \sys with a trusted aggregator with \ac{DP} guarantee is less than $|\lambda_0|$ for \sys when there is no \ac{DP} noise involved.
\end{corollary}

\section{Experimental evaluation}
\label{sec:expts}
%\sayan{needs to be changed with the new experiments}
We implemented \sys and the baseline technique from Section~\ref{sec:PRIEST}.  
We also compare against the results of computing the estimator $\Phi$ (see \Cref{sec:MC}) via \texttt{SecAgg} without adding any \ac{DP} noise.
We refer to this as ``No privacy'' to understand the effect of privacy noise distortion. 
%We implemented the \fulltrust, \aggregator and \distributed models of \sys, displayed in \Cref{alg:sys_model1,alg:sys_model2,alg:sys_model3}, in Python to evaluate their performance.
We performed our evaluation on multiple distributions drawn from the FEMNIST dataset, a widely-used benchmark from the FL community~\cite{Caldas18}.
The theoretical guarantees of \sys are distribution-independent subject to the stated support conditions. 
We therefore use FEMNIST as a controlled federated benchmark to test the finite-sample behavior predicted by the analysis: dependence on the privacy budget $\varepsilon$, estimator parameter $\lambda$, sample count $m$, client-batch size, and divergence magnitude. The objective of these experiments is thus validation of the estimator's theoretical and systems trade-offs.
%Since our primary contribution is the rigorous theoretical bounding of the \sys estimator, this empirical study is  designed to validate these theoretical properties (e.g., the optimal variance parameter $\lambda$) in a canonical federated setting, rather than to exhaustively test generalization across disparate domains.
The data is a collection of images of handwritten digits 0-9, where the image data is partitioned based on the writer of the digit.
In all experiments, clients are partitioned into disjoint batches, as specified in Algorithm~\ref{alg:sys_tagg}.
%Their images are then added to 10 separate stores, each containing images labeled as one of the digits 0-9.
To generate a meaningful distribution for each digit, each $28 \times 28$ image is partitioned into sixteen $7 \times 7$ sub-images, with the mean intensity of each sub-image computed and mapped to a binary level, and the probabilities of encountering each of the $2^{16}$ possibilities listed.
In total there are over 400,000 digits from over 3,500 writers. 
We primarily evaluate the accuracy of the result using the (mean) absolute error (AE) between the ground truth KL divergence and that found by the different estimators. 
The mean is taken over different repetitions of the test with different random choices, and over different distributions induced by pairs of digits. 
By default, we set $\varepsilon = 1.0$ and $\delta = 10^{-6}$. 
Experiments are run in a Jupyter notebook on a commodity laptop with Intel i7-1260P processor and 16GB memory. 
See \texttt{\url{https://figshare.com/s/cbb3d58ad7e42785405d}} for code. 
%Note that in practice, for a 1400-image sample, a subset of these possibilities would typically occur.
%Thereafter, it is simple to use these probabilities to compute the KL divergence between the estimated distributions of any non-identical pair of digits.
%Concerning these pairs, the evaluations will relate to the statistics associated with:
%\begin{itemize}
%    \item the \emph{mean} measured across all $90$ distinct pairs,
%    \item the pair with the \emph{lowest} KL divergence (min pair),
%    \item and the pair with the \emph{highest} KL divergence (max pair).
%\end{itemize}

%\begin{figure*}[tp]
%\centering
%\subfloat[$\varepsilon = 0.05$]{\includegraphics[width=0.3\linewidth]{Eps_exp1_est_a_0.05.png}}
%\qquad
%\subfloat[$\varepsilon = 0.5$]{\includegraphics[width=0.3\linewidth]{Eps_exp1_est_a_0.5.png}}
%\qquad
%\subfloat[$\varepsilon = 3$]{\includegraphics[width=0.3\linewidth]{Eps_exp1_est_a_3.png}}\\
%\caption{\small\label{fig:exp1epsa} Effect of $\lambda$ on mean MSE of \sys models, fixing $|C_{t}\hspace{-0.07em}| \hspace{0.1em} = 200$.}
%\end{figure*}

\subsection{Choice of $\lambda$}
\label{sec:ldachoices}

Our first experiments study the impact of the parameter $\lambda$ on the performance of \sys, complementing the theoretical analysis of the optimal $\lambda$ presented above.
%in \Cref{sec:opt_lambda} of the supplementary material.
Recall that $\lambda$ is used to control the variance of the estimator, and we seek a choice of $\lambda$ that will reliably achieve a small error across a range of scenarios. 
The plots in Figure~\ref{fig:lamdaeps} show results across a choice of values of privacy parameter $\varepsilon$ while sampling clients into batches of 5\% of the total population (error bands show the standard deviation over 10 repetitions).
We see that the estimator with no privacy is able to obtain a consistently good average error close to 0.2.  
The values of KL divergence on this data range between 0.25 to 2.0, meaning that this accuracy is sufficient to accurately distinguish small from large values. 

We use this experiment to determine how best to set the parameter $\lambda$. 
For \sys, the best choice of $\lambda$ is always close to 0, consistent with our analysis in 
Section~\ref{sec:opt_lambda}.
%Theorem~\ref{th:SensitivityOfSampledKLD}, 
%Theorem~\ref{th:gen_estimator_variance}.
%which indicates that the sensitivity (and hence noise) is minimized by this choice. 
At this choice of $\lambda$, 
the observed empirical absolute error is close to that of the estimator without privacy. 
Meanwhile, the baseline approach of adding noise directly to the data is optimized by a value of $\lambda$ below 0, and achieves a lower level of accuracy. 

%We see that for the highest level of privacy ($\varepsilon = 0.1$, Figure~\ref{fig:lambda0.1}), \sys achieves a reduced level of accuracy.
%But for lower privacy requirements the absolute error is closer to that of the estimator without privacy, so that for $\varepsilon \ge 1$ (Figure~\ref{fig:lambda1.0}), there is minimal difference in accuracy. 
%The dependence on $\lambda$ is weak, with good accuracy for all values in the range $0$ to $1$.
%By contrast, the baseline approach is highly sensitive to the choice of $\lambda$, and looses accuracy considerably except at the minima, which occurs for a value of $\lambda$ close to zero. 

A similar pattern is seen in \Cref{fig:Ct}, where we vary $\lambda$, but this time hold $\varepsilon$ constant at $0.5$, and instead vary the number of clients in each batch, from 2\% to 10\%. 
For \sys, the best result is consistently obtained at $\lambda = 0$, where it is close to that with no privacy noise.  
The main effect of varying the batch size is in how quickly the accuracy decreases as $\lambda$ is varied away from 0. 
Meanwhile, the baseline approach achieves weaker accuracy and is highly dependent on the choice of $\lambda$.
Hereafter, we use a default value of $\lambda = 0$ for \sys, since this appears to be a preferred choice, and $\lambda=-1$ for the baseline approach.

\subsection{Performance of \sys models}
\label{sec:sysperf}

Next, we vary the number of samples $m$ used to build each individual estimator. 
Figure~\ref{fig:m} shows results for different choices of $\varepsilon$. 
We see that there is relatively weak dependency on this parameter: although error decreases slowly as $m$ increases, choosing $m=10$ is sufficient to obtain close to the best accuracy. 
%This means that the communication per client is very small: they just need to provide counts for a handful of sampled indices.  
%That is, the per-client communication is less than 1KB, making \sys suitable for very lightweight clients. 
Thus, the application-level contribution from each client consists only of counts for a small number of sampled indices. For $m=10$, this payload is easily below 1\,KB, making \sys suitable for lightweight clients. 
%This figure excludes implementation-dependent cryptographic and protocol overheads introduced by the chosen secure-aggregation mechanism.
Finally, we consider how the accuracy performs for different inputs. 
We consider the KL divergences induced by three pairs of digits: the pair with the lowest KL divergence (3 and 8) of 0.24; the median KL divergence (4 and 1) of 0.76; and the maximum divergence (9 and 0) of 2.0.
The results for these are shown in Figure~\ref{fig:pair} as we vary $\varepsilon$ from 0.1 to 1.5. 

Across all these cases, we see that \sys obtains accuracy that is close to that of the non-private estimator when $\varepsilon$ is moderate in value (0.5 to 1), whereas the baseline approach only becomes competitive for the largest KL divergences when $\varepsilon$ is large. 
We also observe that the error is lower for smaller divergences.  
In the case of the closest pair, the absolute error is close to zero. 
This means that the estimators can accurately distinguish between small and large divergences, with error that appears to scale with the magnitude.  
Taken together, these experiments test \sys across orthogonal sources of difficulty (privacy level, sampling budget, client participation, and distribution separation) within a fixed benchmark, allowing the effects predicted by the analysis to be isolated.

\paragraph{Experimental summary.}
The empirical study  confirms that \sys achieves a practical tradeoff between privacy and accuracy, and is comparable to its non-private version. 
Concretely, we recommend setting the variance parameter $\lambda = 0$ and number of samples to $m=10$, since these choices are seen to be robust in practice. 

\section{Concluding Remarks}\label{sec:concs}
%\sayan{@Graham: pleae review}
\sys, to our knowledge, is the first method for estimating the KL divergence between a public reference distribution and a private, federated data distribution under formal differential privacy guarantees. Our estimator is batch-wise unbiased with provably bounded sensitivity and variance, and its per-client payload is independent of the domain size, remaining below 1\,KB even for domains with $2^{16}$ elements. 
Experiments on FEMNIST show that \textsc{FedPriKL} matches the accuracy of its non-private counterpart at moderate privacy budgets ($\varepsilon \in [0.5, 1]$) and substantially outperforms a baseline in which clients perturb their reports locally, providing accurate estimates to inform distribution-drift and retraining decision. 
%reliably detecting the distribution drifts that trigger model retraining in federated pipelines. 
Future directions include removing the trusted aggregator in favor of fully distributed computation, extending the estimator to broader divergence families, supporting continual drift monitoring over data streams, and making the core protocol robust against malicious participants.

\section*{Acknowledgments}
GC and CM are supported in part by EPSRC
grant EP/V044621/1
The work of GC is supported in part by the Oxford-UBS Centre for Applied AI, sponsored by UBS Group AG.
\newpage
\bibliographystyle{IEEEtran}
\bibliography{IEEEabrv,main}

\begin{thebibliography}{10}
\providecommand{\url}[1]{#1}
\csname url@samestyle\endcsname
\providecommand{\newblock}{\relax}
\providecommand{\bibinfo}[2]{#2}
\providecommand{\BIBentrySTDinterwordspacing}{\spaceskip=0pt\relax}
\providecommand{\BIBentryALTinterwordstretchfactor}{4}
\providecommand{\BIBentryALTinterwordspacing}{\spaceskip=\fontdimen2\font plus
\BIBentryALTinterwordstretchfactor\fontdimen3\font minus \fontdimen4\font\relax}
\providecommand{\BIBforeignlanguage}[2]{{%
\expandafter\ifx\csname l@#1\endcsname\relax
\typeout{** WARNING: IEEEtran.bst: No hyphenation pattern has been}%
\typeout{** loaded for the language `#1'. Using the pattern for}%
\typeout{** the default language instead.}%
\else
\language=\csname l@#1\endcsname
\fi
#2}}
\providecommand{\BIBdecl}{\relax}
\BIBdecl

\bibitem{McCarthy04}
J.~F. {McCarthy \textit{et al.}}, ``Applications of machine learning and high-dimensional visualization in cancer detection, diagnosis, and management,'' \emph{Ann. NY Acad. Sci.}, vol. 1020, no.~1, pp. 239--262, 2004.

\bibitem{Najafabadi15}
M.~M. {Najafabadi \textit{et al.}}, ``Deep learning applications and challenges in big data analytics,'' \emph{J. Big Data}, vol.~2, pp. 1--21, 2015.

\bibitem{Kairouz21Survey}
P.~{Kairouz \textit{et al.}}, ``Advances and open problems in federated learning,'' \emph{Foundations and Trends in ML}, vol.~14, no. 1-2, pp. 1--210, 2021.

\bibitem{Bharadwaj22}
A.~Bharadwaj and G.~Cormode, ``An introduction to federated computation,'' in \emph{ACM SIGMOD}, 2022, pp. 2448--2451.

\bibitem{Abadi16}
M.~{Abadi \textit{et al.}}, ``Deep learning with differential privacy,'' in \emph{ACM CCS}, 2016.

\bibitem{Bonawitz17}
K.~{Bonawitz \textit{et al.}}, ``Practical secure aggregation for privacy-preserving machine learning,'' in \emph{ACM CCS}, 2017.

\bibitem{Kullback51}
S.~Kullback and R.~A. Leibler, ``On information and sufficiency,'' \emph{The Annals of Mathematical Statistics}, vol.~22, no.~1, pp. 79--86, 1951.

\bibitem{DworkRoth14}
C.~Dwork and A.~Roth, ``Algorithmic foundations of differential privacy.'' \emph{Foundations and Trends in TCS}, vol.~9, no. 3-4, pp. 211--407, 2014.

\bibitem{Guha07}
S.~Guha, P.~Indyk, and A.~McGregor, ``Sketching information divergences,'' in \emph{COLT}, 2007, pp. 424--438.

\bibitem{McMahan17}
B.~{McMahan \textit{et al.}}, ``Communication-efficient learning of deep networks from decentralized data,'' in \emph{Artificial Intelligence and Statistics}, 2017, pp. 1273--1282.

\bibitem{DworkDP1}
C.~Dwork, F.~McSherry, K.~Nissim, and A.~Smith, ``Calibrating noise to sensitivity in private data analysis,'' in \emph{Theory of Cryptography}, 2006.

\bibitem{Kenthapadi13}
K.~{Kenthapadi \textit{et al.}}, ``Privacy via the {Johnson-Lindenstrauss} transform,'' \emph{J. Priv. Confidentiality}, vol.~5, no.~1, pp. 39--71, 2013.

\bibitem{Blocki12}
J.~Blocki, A.~Blum, A.~Datta, and O.~Sheffet, ``The {Johnson-Lindenstrauss} transform itself preserves differential privacy,'' in \emph{IEEE FOCS}, 2012.

\bibitem{Hou23}
C.~{Hou \textit{et al.}}, ``Privately customizing prefinetuning to better match user data in federated learning,'' arXiv:2302.09042, 2023.

\bibitem{Nielsen17}
F.~Nielsen, K.~Sun, and S.~Marchand-Maillet, ``On {H{\"o}lder} projective divergences,'' \emph{Entropy}, vol.~19, no.~3, p. 122, 2017.

\bibitem{Andoni09}
A.~Andoni, K.~D. Ba, P.~Indyk, and D.~Woodruff, ``Efficient sketches for earth-mover distance, with applications,'' in \emph{IEEE FOCS}, 2009.

\bibitem{Heikkila19}
M.~Heikkil{\"a}, J.~J{\"a}lk{\"o}, O.~Dikmen, and A.~Honkela, ``Differentially private markov chain monte carlo,'' \emph{NeurIPS}, vol.~32, pp. 1--11, 2019.

\bibitem{Heikkila17}
M.~Heikkil{\"a}, E.~Lagerspetz, S.~Kaski, K.~Shimizu, S.~Tarkoma, and A.~Honkela, ``Differentially private {Bayesian} learning on distributed data,'' \emph{NeurIPS}, vol.~30, pp. 1--10, 2017.

\bibitem{Keeler23}
D.~Keeler, C.~Komlo, E.~Lepert, S.~Veitch, and X.~He, ``{DPrio}: Efficient differential privacy with high utility for {Prio},'' in \emph{PoPETS}, 2023.

\bibitem{Corrigan17}
H.~Corrigan-Gibbs and D.~Boneh, ``Prio: Private, robust, and scalable computation of aggregate statistics,'' in \emph{NSDI}, 2017, pp. 259--282.

\bibitem{Addanki22}
S.~Addanki, K.~Garbe, E.~Jaffe, R.~Ostrovsky, and A.~Polychroniadou, ``Prio+: Privacy preserving aggregate statistics via boolean shares,'' in \emph{Security and Cryptography for Networks}, 2022, pp. 516--539.

\bibitem{FLP}
K.~Bonawitz, P.~Kairouz, B.~McMahan, and D.~R. Ramage, ``Federated learning and privacy,'' \emph{ACM Queue}, vol.~19, 2021.

\bibitem{Kim23}
J.~Kim, G.~Park, M.~Kim, and S.~Park, ``Cluster-based secure aggregation for federated learning,'' \emph{Electronics}, vol.~12, no.~4, p. 870, 2023.

\bibitem{Shamir79}
A.~Shamir, ``How to share a secret,'' \emph{CACM}, vol.~22, no.~11, pp. 612--613, 1979.

\bibitem{Shapiro03}
A.~Shapiro, ``Monte carlo sampling methods,'' \emph{Handbooks in Operations Research and Management Science}, vol.~10, pp. 353--425, 2003.

\bibitem{Amari09}
S.-I. Amari, ``$\alpha$-divergence is unique, belonging to both $f$-divergence and {Bregman} divergence classes,'' \emph{IEEE Trans. Inf. Theory}, vol.~55, no.~11, pp. 4925--4931, 2009.

\bibitem{Sabt_TEE_2015}
M.~Sabt, M.~Achemlal, and A.~Bouabdallah, ``Trusted execution environment: What it is, and what it is not,'' in \emph{IEEE TRUSTCOM}, 2015.

\bibitem{Shepherd:Markantonakis:24}
C.~Shepherd and K.~Markantonakis, \emph{Trusted Execution Environments}.\hskip 1em plus 0.5em minus 0.4em\relax Springer, 2024.

\bibitem{Cramer:Damgard:Nielsen:15}
R.~Cramer, I.~Damg{\aa}rd, and J.~B. Nielsen, \emph{Secure Multiparty Computation and Secret Sharing}.\hskip 1em plus 0.5em minus 0.4em\relax Cambridge University Press, 2015.

\bibitem{AlyS19}
A.~Aly and N.~P. Smart, ``Benchmarking privacy preserving scientific operations,'' in \emph{Applied Cryptography and Network Security}, 2019.

\bibitem{Gentry_FHE_2009}
C.~Gentry, ``Fully homomorphic encryption using ideal lattices,'' in \emph{ACM STOC}, 2009, p. 169–178.

\bibitem{Caldas18}
S.~{Caldas \textit {et al.}}, ``Leaf: A benchmark for federated settings,'' arXiv:1812.01097, 2018.

\end{thebibliography}
%%%%%%%%%%%%%%%%%%%%%%%%%%%%%%%%%%%%%%%%%%%%%%%%%%%%%%%%%%%%

\clearpage

\appendices

\section*{ADDITIONAL PROOFS}\label{app:proofs}

This section of provides detailed proofs of the results omitted from the main paper.

\begin{comment}
\subsection{Proof of \Cref{th:Model3_DP}}
\begin{proof}
Under our notion of adjacency, which corresponds to example-level \ac{DP}, making one change to the input can change $P(x_{i})$ by at most $1/N$, where $N$ is the number of samples used to estimate $P(x_{i})$.
Moreover, looking at $P(x_{i})$ for a collection of samples $x_{i}$ can be considered as a histogram query, so we can add total noise with sensitivity $2/N$ to each sampled item.
Consequently, we obtain an $(\varepsilon, \delta)$-DP guarantee by adding the Gaussian noise as specified in Algorithm 1.
\end{proof}
\end{comment}

\subsection{Proof of \Cref{th:SensitivityOfKLD}}
For distributions $P_{1}, P_{2} \in \mathcal{P}^{>1}$ such that $P_{1} \sim P_{2}$, there must be histograms
$h_{1}, h_{2} \in \mathcal{H}^{>1}_{\mathcal{X}}$ with $h_{1} \sim h_{2}$ satisfying $P_{1} = \psi(h_{1})$ and $P_{2} = \psi(h_{2})$.
Thus, there must be some $x_{1}, x_{2} \in \mathcal{X}$ such that
$h_{1}(x_{1}) - h_{2}(x_{1}) = 1$, $h_{1}(x_{2}) - h_{2}(x_{2}) = -1$ and $h_{1}(x) = h_{2}(x) \,\forall \, x \in \mathcal{X} \setminus \{x_{1}, x_{2} \}$.
Let $N = |h_{1}| = |h_{2}|$, $n_{1} = h_{1}(x_{1})$, and $n_{2} = h_{1}(x_{2})$.
Because $h_{1}(x) = h_{2}(x)$ for $x \in \mathcal{X} \setminus \{x_{1}, x_{2} \}$, let the value of $h_{1}(x_{1}) + h_{1}(x_{2}) = h_{2}(x_{1}) + h_{2}(x_{2})$ be denoted by $k$.
Note that $P_{1}, P_{2} \in \mathcal{P}^{>1}$ implies that $k \geq 4$ (because each value from $\mathcal{X}$ occurs at least twice).
Therefore, we get
\begin{align}
    &|f(P_{1}) - f(P_{2})|= \left \lvert \sum_{x} \Pi     (x) \ln{ \frac{ \Pi     (x)}{P_{1}(x)}}-\sum_{x} \Pi     (x) \ln{\frac{\Pi     (x)}{P_{2}(x)}} \right \rvert \nonumber \\
    &= \left \lvert \sum_{x} \Pi     (x) \ln{\frac{P_{2}(x)}{P_{1}(x)}} \right \rvert \nonumber\\
    &= \left \lvert     \Pi     (x_{1}) \ln{\frac{h_{2}(x_{1})}{h_{1}(x_{1})}} + \Pi     (x_{2}) \ln{\frac{h_{2}(x_{2})}{h_{1}(x_{2})}} \right \rvert \nonumber\\ 
    &= \left \lvert     \Pi     (x_{1}) \ln{\frac{n_{1}-1}{n_{1}}} + \Pi     (x_{2}) \ln{\frac{n_{2} + 1}{n_{2}}} \right \rvert \nonumber \\
    &= \left \lvert     \Pi     (x_{1}) \ln{\left(1 - \frac{1}{n_{1}} \right)} + \Pi     (x_{2}) \ln{\left(1 + \frac{1}{k - n_1} \right)} \right \rvert \nonumber \\
    &\leq \left \lvert     \Pi (x_{1}) \ln{\left(1 - \frac{1}{n_{1}} \right)} \right \rvert + \left \lvert     \Pi     (x_2) \ln{\left(1 + \frac{1}{k - n_{1}} \right)} \right \rvert \nonumber\\
    & = \Pi     (x_{1}) \ln{\left(1 + \frac{1}{n_{1} - 1} \right)} + \Pi     (x_{2}) \ln{\left(1 + \frac{1}{k - n_{1}} \right)} \nonumber \\
    &\leq \Pi     (x_{1}) \ln{\left(1 + \frac{1}{n_{1} - 1}\right)} \nonumber\\
    &+ \Pi     (x_{2}) \ln{\left(1 + \frac{1}{(n_{1} + 2) - n_{1}} \right)}\nonumber\\ 
    &\text{[as $k \geq n_{1} + 2$ \& $\ln{\left(1 + \frac{1}{k - n_{1}} \right)}$ decreases with $k$]} \nonumber \\
    &\leq \Pi     (x_{1}) \ln{\left(1 + \frac{1}{n_{1} - 1} \right)} + \Pi     (x_{2}) \ln{\frac{3}{2}}\nonumber\\
    &\leq \Pi     (x_{1}) \ln{\left(1 + \frac{1}{2 - 1} \right)} + \Pi     (x_{2}) \ln{\frac{3}{2}} \nonumber \\
    &\text{[as $n_{1} \geq 2$ \& $\ln{\left(1 + \frac{1}{n_{1} - 1} \right)}$ decreases with $n_{1}$]}\nonumber\\
    &\leq \ln{2}(\Pi     (x_{1}) + \Pi     (x_{2})) \leq \ln{2} < 0.7. \nonumber
\end{align}
\qed

\subsection{Proof of \Cref{th:SensitivityOfSampledKLD}} \label{app_sec:sensitivity_estimate}
    In order to prove this result, we derive the proof of a more generic result where the estimation of the KL divergence between $\Pi$ and $P_i$ uses the variance parameter $\lambda_{P_i}$ for $i=1,2$ such that $\lambda_{P_1}$ and $\lambda_{P_2}$ may or may not be the same. In the special case where $\lambda_{P_1}=\lambda_{P_2}=\lambda$, we recover the setting specified in the theorem.
    
    For $P_{1}, P_{2} \in \mathcal{P}^{>1}$ with $P_{1} \sim P_{2}$, we have $h_{1}, h_{2} \in \mathcal{H}^{>1}_{\mathcal{X}}$ with $h_{1} \sim h_{2}$ satisfying $P_{1} = \psi(h_{1})$ and $P_{2} = \psi(h_{2})$.
    Thus, there are $x_{1}, x_{2} \in \mathcal{X}$ such that
    $h_{1}(x_{1}) - h_{2}(x_{1}) = 1$, $h_{1}(x_{2}) - h_{2}(x_2) = - 1$ and $h_{1}(x) = h_{2}(x) \,\forall \, x \in \mathcal{X} \setminus\{ x_1, x_2 \}$.
    Let $N = |h_{1}| = |h_{2}|$, $n_{1} = h_{1}(x_{1})$, and $n_{2} = h_{1}(x_{2})$.
    Note that, as $P_{1}, P_{2} \in \mathcal{P}^{>1}$, $n_{1}, n_{2} \geq 2$.
    Let $X = x \in \mathcal{X}$ be some realization of $X \sim \Pi$, and, for notational convenience, let $r_{i} = \frac{P_{i}(x)}{\Pi(x)}$ and $\lambda_{i} = \lambda_{P_{i}}$ for $i = 1, 2$ in the context of this proof.
    We obtain $|\hat{f}(P_{1}) - \hat{f}(P_{2})|$
    \begin{align*}
        &= |(\lambda_{1} (r_{1} - 1) - \ln{r_{1}}) - (\lambda_{2}(r_{2} - 1) - \ln{r_{2}}) | \nonumber\\
        &= \left \lvert \frac{ \lambda_{1} P_{1}(x)-\lambda_{2} P_{2}(x)}{\Pi(x)} + (\lambda_{1} - \lambda_{2}) + \ln{\frac{P_{2}(x)}{P_{1}(x)}} \right \rvert \nonumber \\
        &\leq \left \lvert \frac{ \lambda_{1} P_{1}(x)-\lambda_{2} P_{2}(x)}{\Pi(x)} \right \rvert + \lvert \lambda_{1} - \lambda_{2} \rvert + \left| \ln{ \frac{P_{2}(x)}{P_{1}(x)}}   \right| \nonumber\\
        &[\text{by triangle-inequality}] \nonumber \\
        &= \left \lvert \frac{ \lambda_{1} P_{1}(x)-\lambda_{2} P_{2}(x)}{\Pi(x)} \right \rvert + \lvert \lambda_{1} - \lambda_{2} \rvert\nonumber\\
        &+ \left| \ln{ \frac{h_{2}(x)}{h_{1}(x)}} \right| \left[ \text{as } P_{i}(x) = \frac{h_{i}(x)}{N} \,\text{for} \, i = 1, 2 \right] \nonumber \\
        &\leq \left \lvert \frac{ \lambda_{1} P_{1}(x) - \lambda_{2} P_{2}(x)}{\Pi(x)} \right \rvert + \lvert \lambda_{1} - \lambda_{2} \rvert\nonumber\\
        &+ \max \left\{ \left| \ln{ \frac{n_{1}}{n_{1} - 1}} \right|, \left|\ln{ \frac{n_{2} + 1}{n_{2}}} \right| \right\} \nonumber\\
        &\left[\text{as }h_{1}(x) = h_{2}(x)\forall x \neq x_{1}, x_{2},\text{ or } \ln{ \frac{h_{2}(x)}{h_{1}(x)}} = 0 \right] \nonumber \\
        &\leq \max \left\{ \frac{p | \lambda_{1} - \lambda_{2}|}{\Pi(x)}, \frac{|\lambda_{1} n_{1} - \lambda_{2}(n_{1} - 1)|}{\Pi(x_{1})},   \frac{|\lambda_{1} n_{2} - \lambda_{2} (n_{2} + 1)|}{N\Pi(x_2)} \right\} \nonumber 
        \\         
        &+ \lvert \lambda_{1} - \lambda_{2} \rvert + \max \left\{ \ln{ \frac{n_{1}}{n_{1} - 1}},  \ln{\frac{n_{2} + 1}{n_{2}}} \right\}\nonumber
        \\ 
        &\left[ \text{ where } p = h_{1}(x) / N = h_{2}(x) / N \forall x \in \mathcal{X} \setminus\{ x_{1}, x_{2} \} \right] \nonumber \\
        &\leq \max \left\{ \hat{\alpha} |\lambda_{1} - \lambda_{2}|,  \frac{|n_{1}(\lambda_{1} - \lambda_{2}) + \lambda_{2}|}{N \Pi(x_{1})},   \frac{|n_{2}(\lambda_{1} -\lambda_{2}) - \lambda_{2}|}{N \Pi(x_{2})} \right\} \\\nonumber 
        &+ \lvert \lambda_{1} - \lambda_{2} \rvert + \max \left\{ \ln{ \frac{n_{1}}{n_{1} - 1}},   \ln{\frac{n_{2} + 1}{n_{2}}} \right\} \nonumber \\
        &\leq \max \left\{ \hat{\alpha} |\lambda_{1} - \lambda_{2}|,   \frac{|n_{1}(\lambda_{1} - \lambda_{2})|   +   |\lambda_{2}|}{N \Pi(x_1)},   \frac{|n_{2}(\lambda_{1} -\lambda_{2})|   +   |\lambda_{2}|}{N \Pi(x_{2})} \right\}\nonumber\\
        &+\lvert \lambda_{1} - \lambda_{2} \rvert + \max \left\{ \ln{ \frac{n_{1}}{n_{1} - 1}},   \ln{\frac{n_{2} + 1}{n_{2}}} \right\} \nonumber \\
        &\leq \max \left\{ \hat{\alpha} |\lambda_{1} - \lambda_{2}|,   \hat{\alpha}|\lambda_{1} - \lambda_{2}|   +   \frac{|\lambda_{2}|}{N \Pi(x_1)}, \right.\nonumber\\
        &\left.  \hat{\alpha}|\lambda_{1} - \lambda_{2}|   +   \frac{|\lambda_{2}|}{N \Pi(x_2)} \right\} 
        + \lvert \lambda_{1} - \lambda_{2} \rvert \nonumber\\
        &+ \max \left\{ \ln{ \frac{n_{1}}{n_{1} - 1}},   \ln{\frac{n_{2} + 1}{n_{2}}} \right\} \nonumber \\
        &= (\hat{\alpha} + 1) |\lambda_{1} - \lambda_{2}|   +   \frac{|\lambda_{2}|}{N} \max \left\{ \frac{1}{\Pi(x_{1})},   \frac{1}{\Pi(x_{2})} \right\} \nonumber\\
        &+ \max \left\{ \ln{ \frac{n_{1}}{n_{1} - 1}},   \ln{\frac{n_{2} + 1}{n_{2}}} \right\} \nonumber \\
        &\leq (\hat{\alpha} + 1)|\lambda_{1} - \lambda_{2}|   +   \frac{|\lambda_{2}|}{N\pi_{\min}} \nonumber\\
        &+ \max \left\{ \ln{ \frac{n_{1}}{n_{1} - 1}},   \ln{\frac{n_{2} + 1}{n_{2}}} \right\} \nonumber \\
         &\leq(\hat{\alpha} + 1)|\lambda_{1} - \lambda_{2}|   +   \frac{|\lambda_{2}|}{N\pi_{\min}}\nonumber\\ 
         &+ \max \left\{ \ln{ \left(1 + \frac{1}{n_{1} - 1} \right)},  \ln{\left(1 + \frac{1}{n_{2}} \right)} \right\} \nonumber \\
         &\leq(\hat{\alpha} + 1)|\lambda_{1} - \lambda_{2}|   +   \frac{|\lambda_{2}|}{N\pi_{\min}}\nonumber\\ 
         &+ \max \left\{ \ln{ \left( 1 + \frac{1}{2 - 1}\right)},   \ln{ \left(1 + \frac{1}{2} \right)} \right\} \nonumber \\
         &= (\hat{\alpha} + 1)|\lambda_{1} - \lambda_{2}|   +   \frac{|\lambda_{2}|}{N\pi_{\min}} + \ln{2}. \nonumber
    \end{align*}
    Substituting $\lambda$ for $\lambda_1$ and $\lambda_2$ gives us the desired result.  
\qed

\subsection{Proof of \Cref{th:Model2_DP}}
    By \Cref{th:SensitivityOfSampledKLD}, for any $x\in\mathcal{X}$ and a chosen $\lambda$, the sensitivity of $\Phi_{\Pi,P}(r(x),\lambda)$, %denoted by $\Delta\left(\hat{D}_{\operatorname{KL}}(\Pi\| P)\right)$,
    is bounded above by $\frac{|\lambda|}{\pi_{\min}m} + \ln{2}$. As this upper bound is independent of $T$, let us denote this term as $\kappa$, a constant (w.r.t. $T$). Therefore, for \emph{each} set of clients $C_t$ in any batch $t$, for $m$ instances of $x$ sampled from $\Pi$, the total sensitivity of  $\sum_{i_{t}=1}^m\Phi_{\Pi,P}(r(x_{i_t}),\lambda)$ is independently bounded above by $m\kappa$. Moreover, as $C_1,\ldots, C_T$ are all disjoint, $\sum_{t=1}^T\sum_{i_{t}=1}^m\Phi_{\Pi,P}(r(x_{i_t}),\lambda)$ is formally equivalent to computing $\sum_{i_{t}=1}^m\Phi_{\Pi,P}(r(x_{i_t}),\lambda)$ in $T$ independent and parallel rounds over the disjoint sets of clients $C_1,\ldots,C_T$ and obtaining the final sum over $i_1,\ldots,i_T$ . Hence, by the parallel compositionality property of DP, the total sensitivity of $\sum_{t=1}^T\sum_{i_{t}=1}^m\Phi_{\Pi,P}(r(x_{i_t}),\lambda)$ is bounded above by $m\kappa$. Thus, in order to show that the final estimator given by Algorithm 2 satisfies $(\varepsilon,\delta)$-DP, by the definition of DP with additive Gaussian noise, it suffices to show $\eta_{\epsilon,\delta}\sim \mathcal{N}\left(0,\frac{\ln{(1.25/\delta)} \Delta\left(\frac{1}{mT}\sum_{t=1}^T\sum_{i_{t}=1}^m\Phi_{\Pi,P}(r(x_{i_t}),\lambda)\right)^2}{\epsilon^2}\right)$. But we obtain
    \begin{align}
    &\Delta\left(\frac{1}{mT}\sum_{t=1}^T\sum_{i_{t}=1}^m\Phi_{\Pi,P}(r(x_{i_t}),\lambda)\right)\nonumber\\
    &=\frac{1}{mT}\Delta\left(\sum_{t=1}^T\sum_{i_{t}=1}^m\Phi_{\Pi,P}(r(x_{i_t}),\lambda)\right)\nonumber\\
    &=\frac{1}{mT}m\kappa=\frac{1}{T}\Delta\left(\hat{D}_{\operatorname{KL}}(\Pi\| P)_{[\lambda,T]}\right).\label{eq:sensitivity_total}
    \end{align}
    Hence, recalling that the noise $\eta_{\varepsilon,\delta}\sim \mathcal{N}\left(0,\frac{2\ln(1.25/\delta)\Delta\left(\hat{D}_{\operatorname{KL}}(\Pi\| P)_{[\lambda,T]}\right)^2}{(\varepsilon T)^2}\right)$, and using \eqref{eq:sensitivity_total} in this expression, we obtain $\eta_{\epsilon,\delta}\sim \mathcal{N}\left(0,\frac{\ln{(1.25/\delta)} \Delta\left(\frac{1}{mT}\sum_{t=1}^T\sum_{i_{t}=1}^m\Phi_{\Pi,P}(r(x_{i_t}),\lambda)\right)^2}{\epsilon^2}\right)$, as needed. \qed
    
    %$\frac{\ln{(1.25/\delta)} \Delta\left(\frac{1}{mT}\sum_{t=1}^T\sum_{i_{t}=1}^m\Phi_{\Pi,P}(r(x_{i_t}),\lambda)\right)^2}{\epsilon^2}$, we obtain 

    %The total noise added to the estimator by the trusted aggregator is $\frac{1}{mT}\sum_{i=1}^{mT}\eta_{\varepsilon,\delta}$ where $\eta_{\varepsilon,\delta}\sim \mathcal{N}\left(0,\frac{2\ln(1.25/\delta)\Delta\left(D_{\operatorname{KL}(\Pi\| P)}\right)^2}{\varepsilon^2/mT}\right)$. Therefore, by the property of Gaussian distribution, the total noise added also follows a Gaussian distribution centered at $0$ and having a variance $\operatorname{Var}\left[\frac{1}{mT}\sum_{i=1}^{mT}\eta_{\varepsilon,\delta}\right]=\frac{1}{(mT)^2}\sum_{i=1}^{mT}\operatorname{Var}\left[\eta_{\varepsilon,\delta}\right]=\frac{1}{mT}\operatorname{Var}\left[\eta_{\varepsilon,\delta}\right]= \frac{2\ln(1.25/\delta)\Delta\left(D_{\operatorname{KL}(\Pi\| P)}\right)^2}{\varepsilon^2}$. Consequently, we satisfy $(\varepsilon,\delta)$-\ac{DP}. 

\subsection{Proof of \Cref{th:estimator_unbiased}}

\begin{align*} 
&\mathbb{E} \left[ \hat{D}_{\operatorname{KL}} \left( \Pi \| P \right)_{[\lambda, T, \varepsilon, \delta]} \right] \nonumber\\
&= \mathbb{E}\left[\frac{1}{mT}\sum\limits_{t=1}^{T}\sum\limits_{i_t = 1}^{m}\Phi'_{\Pi,P}(r(x_{i_t}),\lambda)\right]\nonumber\\
&=\mathbb{E}\left[\frac{1}{mT}\sum\limits_{t=1}^{T}\sum\limits_{i_t = 1}^{m}\Phi_{\Pi,P}(r(x_{i_t}),\lambda)\right][\text{since } \mathbb{E} \left[ \eta_{\varepsilon, \delta} \right] = 0] \nonumber \\
&\frac{1}{mT}\sum\limits_{t=1}^{T}\sum\limits_{i_t = 1}^{m}\mathbb{E}\left[\Phi_{\Pi,P}(r(x_{i_t}),\lambda)\right]\nonumber\\
&=\mathbb{E}_{X\sim \Pi}\left[\Phi_{\Pi,P}(r(X),\lambda)\right]
    %&= \mathbb{E} \left[ \frac{1}{T} (\sum_{t = 1}^{T} \lambda_{0}(r(x_{t}) - 1) - \ln{r(x_{t})}) \right] \nonumber \\
    %&= \frac{T}{T} \mathbb{E}_{X \sim \Pi} \left[ \lambda_{0}(r(X) - 1) - \ln{r(X)}) \right] \left(\text{as } x_{1}, \ldots, x_{T} \sim \Pi \right) \nonumber \\
    = D_{\operatorname{KL}}(\Pi \|P)\nonumber\\
    &[\text{as $x_{i_t}$s are i.i.d. and using Facts~\ref{lem:estimator_fraction_expect} and \ref{lem:estimator_sum_expect}} ].
\end{align*}
\qed

\subsection{Proof of \Cref{th:gen_estimator_variance}}
    \begin{align}
        &\operatorname{Var} \left[ \hat{D}_{\operatorname{KL}} \left( \Pi \| P \right)_{[\lambda, T, \varepsilon, \delta]} \right]  = \operatorname{Var}\left[ \frac{1}{mT}\sum\limits_{t=1}^{T}\sum\limits_{i = 1}^{m}\Phi'_{\Pi,P}(r(x_i),\lambda) \right]\nonumber\\
        &=\operatorname{Var}\left[ \frac{1}{mT}\sum\limits_{t=1}^{T}\sum\limits_{i = 1}^{m}\Phi_{\Pi,P}(r(x_i),\lambda) \right]+\frac{1}{(mT)^2}mT\operatorname{Var}\left[\eta_{\varepsilon,\delta}\right]\nonumber\\
        &=\frac{1}{(mT)^2}\sum\limits_{t=1}^{T}\sum\limits_{i = 1}^{m}\operatorname{Var}\left[\Phi_{\Pi,P}(r(x_i),\lambda)\right]+ \frac{1}{mT}\operatorname{Var}\left[\eta_{\varepsilon,\delta}\right]\nonumber\\
        &=\frac{1}{(mT)^2}(mT)\operatorname{Var}_{X\sim \Pi}\left[\Phi_{\Pi,P}(r(X),\lambda)\right]+ \frac{1}{mT}\operatorname{Var}\left[\eta_{\varepsilon,\delta}\right]\nonumber\\
        &[\text{as }x_1,\ldots x_m \overset{\text{i.i.d.}}{\sim}\Pi]\nonumber\\
        &=\frac{1}{mT}\operatorname{Var}_{X\sim \Pi}\left[\Phi_{\Pi,P}(r(X),\lambda)\right]+ \frac{1}{mT}\operatorname{Var}\left[\eta_{\varepsilon,\delta}\right]\nonumber\\
        &=\frac{1}{mT} \left(\lambda^{2} \operatorname{Var}[r(X) - 1] + \operatorname{Var}[\ln{r(X)}]\right.\nonumber\\  &\left. -2\lambda \left(\operatorname{Cov} \left[(r(X) - 1), \ln{r(X)} \right] \right)\right) +\frac{1}{mT}\operatorname{Var}\left[\eta_{\varepsilon,\delta}\right]\nonumber\\
        &=\frac{1}{mT} \left(\lambda^{2} \operatorname{Var}[r(X)] + \operatorname{Var}[\ln{r(X)}]\right.\nonumber\\  &\left. -2\lambda\left(\mathbb{E} \left[(r(X) - 1) \ln{r(X)} \right]-\mathbb{E}[r(X) - 1]  \mathbb{E}[\ln{r(X)}]\right)\right) \nonumber\\
        &+\frac{1}{mT}\operatorname{Var}\left[\eta_{\varepsilon,\delta}\right]\nonumber\\
        &=\frac{1}{mT} \left(\lambda^{2} \operatorname{Var}[r(X)] + \operatorname{Var}[\ln{r(X)}]\right.\nonumber\\ &\left.  -2\lambda\left(\mathbb{E} \left[(r(X) - 1) \ln{r(X)} \right]\right)\right) +\frac{1}{mT}\operatorname{Var}\left[\eta_{\varepsilon,\delta}\right]\nonumber\\
        &[\text{as } \mathbb{E}_{X\sim \Pi}[r(X)-1]=0]\nonumber\\
        &=\frac{1}{mT} \left(\lambda^{2} \operatorname{Var}[r(X)] + \operatorname{Var}[\ln{r(X)}]\right.\nonumber\\  &\left. -2\lambda\left(\mathbb{E} \left[(r(X)\ln{r(X)} \right]\right)+2\lambda\mathbb{E}\left[\ln{r(X)}\right]\right)\nonumber\\ 
        &+\frac{1}{mT}\operatorname{Var}\left[\eta_{\varepsilon,\delta}\right]\nonumber\\
        &= \frac{1}{mT}\left(\lambda^{2} \operatorname{Var}[r(X)] + \operatorname{Var}[\ln{r(X)}]\right.\nonumber\\ &\left.- 2\lambda \sum_{x \in \mathcal{X}} \frac{P(x)}{\Pi(x)} \ln{\frac{P(x)}{\Pi(x)}} \Pi(x) + 2\lambda \sum_{x \in \mathcal{X}} \ln{\frac{P(x)}{\Pi(x)}} \Pi(x)\right)\nonumber\\ &+\frac{1}{mT}\operatorname{Var}\left[\eta_{\varepsilon,\delta}\right]\nonumber \\
        &=\frac{1}{mT}\left(\lambda^{2} \operatorname{Var}[r(X)] + \operatorname{Var}[\ln{r(X)}]\right.\nonumber\\ &\left.- 2\lambda\left(D_{\operatorname{KL}}(P \| \Pi)+ D_{\operatorname{KL}}(\Pi \| P)\right)\right) +\frac{1}{mT}\operatorname{Var}\left[\eta_{\varepsilon,\delta}\right]\nonumber \\
        &=\frac{1}{mT}\left(\lambda^{2} \operatorname{Var}[r(X)] + \mathbb{E}[\ln^2{r(X)}] -\mathbb{E}[\ln{r(X)}]^2\right.\nonumber\\  
        &\left.- 2\lambda\left(D_{\operatorname{KL}}(P \| \Pi)+ D_{\operatorname{KL}}(\Pi \| P)\right)\right) +\frac{1}{mT}\operatorname{Var}\left[\eta_{\varepsilon,\delta}\right]\label{eq:var_bound_1}
        \end{align}
Substituting $\mathbb{E}_{X\sim \Pi}[\ln r(X)]^2$ = $\left(\sum_{x\in \mathcal{X}}\Pi(x)\ln{\frac{P(x)}{\Pi(x)}}\right)^2 =D_{\operatorname{KL}}(\Pi\|P)^2$ into \eqref{eq:var_bound_1}, we obtain:
    \begin{align}
    &=\frac{1}{mT}\left(\lambda^{2} \operatorname{Var}[r(X)] + \mathbb{E}[\ln^2{r(X)}] -D_{\operatorname{KL}}(\Pi\|P)^2\right.\nonumber\\
    &-\left.2\lambda\left(D_{\operatorname{KL}}(P \| \Pi)+ D_{\operatorname{KL}}(\Pi \| P)\right)\right) +\frac{1}{mT}\operatorname{Var}\left[\eta_{\varepsilon,\delta}\right]\nonumber\\
    &=\frac{1}{mT}\left(\lambda^{2} \mathbb{E}[r(X)^2] -\lambda^2\mathbb{E}[r(X)]^2 + \mathbb{E}[\ln^2{r(X)}] \right.\nonumber\\
    &\left.-D_{\operatorname{KL}}(\Pi\|P)^2  - 2\lambda\left(D_{\operatorname{KL}}(P \| \Pi)+ D_{\operatorname{KL}}(\Pi \| P)\right)\right)\nonumber\\ &+\frac{1}{mT}\operatorname{Var}\left[\eta_{\varepsilon,\delta}\right]\nonumber\\
    &=\frac{1}{mT}\left(\lambda^{2} \mathbb{E}[r(X)^2] -\lambda^2 + \mathbb{E}[\ln^2{r(X)}] -D_{\operatorname{KL}}(\Pi\|P)^2\right.\nonumber\\ 
    &-\left.2\lambda\left(D_{\operatorname{KL}}(P \| \Pi)+ D_{\operatorname{KL}}(\Pi \| P)\right)\right) +\frac{1}{mT}\operatorname{Var}\left[\eta_{\varepsilon,\delta}\right]\label{eq:var_bound_2}
    \end{align}
    %Recalling that $1 - \frac{1}{x} \leq \ln{x} \leq x - 1 \,\forall \, x \in \mathbb{R}_{\geq 0}$, we get
    %\[
    %\ln^{2}(r(X))\leq 
    %\begin{cases}
    %    (r(X) - 1)^{2}, \quad \text{ if } r(X) \geq 1\\
    %    \left(1 - \frac{1}{r(X)}\right)^{2}, \quad \text{ if } r \in (0, 1).
    %\end{cases}
    %\]
    %Therefore, $\mathbb{E} \left[ \ln^{2}(r(X)) \right]$
    Using $|\ln x| \le |x-1|/\sqrt{x}$ for all $x>0$ (equivalently, $|u| \le 2\sinh(|u|/2)$ with $u=\ln x$), we get
\[
\ln^2 r(X) \;\le\; \frac{(r(X)-1)^2}{r(X)} \;=\; |r(X)-1|\cdot\left|1-\frac{1}{r(X)}\right| .
\]
Therefore, by the Cauchy-Schwarz inequality,
\begin{align}
    &\mathbb{E}\!\left[\ln^2 r(X)\right]
    \le \sqrt{\mathbb{E}\!\left[(r(X)-1)^2\right]\,\mathbb{E}\!\left[\left(1-\tfrac{1}{r(X)}\right)^2\right]} \nonumber\\
        %&\le \max\left\{\mathbb{E}\!\left[(r(X)-1)^2\right],\ \mathbb{E}\!\left[\left(1-\tfrac{1}{r(X)}\right)^2\right]\right\}\\
        &\leq \max \left\{ \mathbb{E}  \left[(r(X) - 1)^{2} \right], \mathbb{E} \left[ \left(1 - \frac{1}{r(X)} \right)^{2} \right] \right \} \nonumber \\
        &= \max \left\{ \mathbb{E}   \left[ r( X)^{2} \right] + 1 - 2\ \mathbb{E}   \left[ r(X) \right],\right.\nonumber\\
        &\left.1 + \mathbb{E}   \left[ \frac{1}{r( X)^{2}} \right] - 2\ \mathbb{E}   \left[ \frac{1}{r( X)} \right] \right\} \nonumber \\
        &= \max \left\{ \mathbb{E}   \left[ r( X)^{2} \right] - 1,     1 + \mathbb{E}   \left[ \frac{1}{r( X)^{2}} \right] - 2\ \mathbb{E}   \left[ \frac{1}{r( X)} \right] \right\} \nonumber \\
        &\leq \max \left\{ \mathbb{E}   \left[r( X)^{2} \right] - 1,     1 + \mathbb{E}   \left[ \frac{1}{r( X)^{2}} \right] - \frac{2}{\mathbb{E}   \left[r( X) \right]} \right\} \nonumber \\
        &[ \text{as $\mathbb{E}   \left[ \frac{1}{z} \right] \geq \frac{1}{\mathbb{E}   \left[z \right]}\forall z>0$ by Jensen's inequality}] \nonumber \\
        &= \max \left\{ \mathbb{E}   \left[r( X)^{2} \right] - 1,     \mathbb{E}   \left[ \frac{1}{r( X)^{2}} \right] - 1 \right\} \nonumber \\
        &\leq \max \left\{ \sum_{x \in \mathcal{X}} \frac{P(x)^{2}}{\Pi(x)^{2}}     \Pi(x), \sum_{x \in \mathcal{X}} \frac{\Pi(x)^{2}}{P(x)^{2}}     \Pi(x) \right\} - 1 \nonumber \\
        &= \max \left\{ \sum_{x \in \mathcal{X}} \frac{P(x)^2}{\Pi(x)}, \sum_{x \in \mathcal{X}} \frac{\Pi(x)^{3}}{P(x)^{2}} \right\} - 1 \nonumber \\
        &\leq \max \left\{ \sum_{x \in \mathcal{X}} \alpha P(x),\sum_{x \in \mathcal{X}} \beta^{2} \Pi(x) \right \} - 1\nonumber\\
        &= \max\{\alpha,\beta^{2}\} - 1. \label{eq:gen_estimator_variance_calculation2}
    \end{align}
    Moreover, $\mathbb{E}_{X\sim \Pi}[r(X)^2]=\sum_{x\in\mathcal{X}}\Pi(x)\frac{P^2(x)}{\Pi^2(x)}=\sum_{x\in\mathcal{X}}\frac{P^2(x)}{\Pi(x)}\leq \alpha$. Combining this and \eqref{eq:gen_estimator_variance_calculation2} with \eqref{eq:var_bound_2} gives us:
    %\begin{align}
    %&= \sigma_{\varepsilon, \delta}^{2} + \frac{1}{T} \left( \lambda^{2}     \mathbb{E}    \left[ r( X)^{2} \right] - \lambda^{2}     \mathbb{E}^{2}    \left[ r( X) \right] + \mathbb{E}   \left[ \ln^{2}{r(x_t)} \right] \right) \nonumber \\
    %    &\qquad + \frac{1}{T}D_{\operatorname{KL}}(\Pi \| P)^{2} - \frac{2\lambda}{T} (D_{\operatorname{KL}}(P \| \Pi) + D_{\operatorname{KL}}(\Pi \| P)) \nonumber \\
    %    &= \sigma_{\varepsilon, \delta}^{2} + \frac{1}{T} \left( \lambda^{2}     \mathbb{E}    \left[ r( X)^{2} \right]-\lambda^{2} + \mathbb{E}   \left[ \ln^{2}{r( X)} \right] \right) \nonumber \\
    %    &\qquad + \frac{1}{T}D_{\operatorname{KL}}(\Pi \| P)^{2} - \frac{2\lambda}{T} (D_{\operatorname{KL}}(P \| \Pi) + D_{\operatorname{KL}}(\Pi \| P)) \nonumber \\
    %    &\left[ \because \text{\Cref{lem:estimator_fraction_expect}} \implies \mathbb{E}[r( X)] = 1 \right] \nonumber \\
     %   &= \sigma_{\varepsilon, \delta}^{2} + \frac{1}{T} \left( \lambda^{2}     \mathbb{E}    \left[ r( X)^{2} \right] + \mathbb{E}   \left[ \ln^{2} {r( X)} \right] \right) - \frac{\lambda^{2}}{T} \nonumber \\
      %  &\qquad + \frac{1}{T}D_{\operatorname{KL}}(\Pi \| P)^{2} - \frac{2\lambda}{T} (D_{\operatorname{KL}}(P \| \Pi) + D_{\operatorname{KL}}(\Pi \| P)). \label{eq:gen_estimator_variance_calculation1}
    %\end{align}

    %Combining \eqref{eq:gen_estimator_variance_calculation1} and \eqref{eq:gen_estimator_variance_calculation2} gives us:
    \begin{align}
        &\operatorname{Var} \left[ \hat{D}_{\operatorname{KL}} \left( \Pi \| P \right)_{[\lambda, T, \varepsilon, \delta]} \right] \nonumber \\
        &\leq \frac{1}{mT}\left(\lambda^{2} (\alpha -1) + \max\{\alpha,\beta^2\} -1 -D_{\operatorname{KL}}(\Pi\|P)^2\right.\nonumber\\
        &\left.- 2\lambda\left(D_{\operatorname{KL}}(P \| \Pi)+ D_{\operatorname{KL}}(\Pi \| P)\right)+\operatorname{Var}[\eta_{\varepsilon,\delta}]\right) \label{eq:var_bound_3}
    \end{align}
    Recalling the definition of $\eta_{\varepsilon,\delta}$ in \Cref{alg:sys_tagg}, we have
\begin{equation}\label{eq:var_bound_4}
\mathrm{Var}[\eta_{\varepsilon,\delta}] = \frac{2\ln(1.25/\delta)\, m \left(\frac{|\lambda|}{N\pi_{\min}} + \ln 2\right)^2}{\varepsilon^2}. \tag{6}
\end{equation}
    %Recalling the definition of $\eta_{\varepsilon,\delta}$ and using Theorem 4.5, we get
    %\begin{align}
    %&\operatorname{Var}[\eta_{\varepsilon,\delta}]\leq\frac{2\ln(1.25/\delta)\left(\frac{|\lambda|}{N\pi_{\min}}+\ln{2}\right)^2}{(\varepsilon T)^2}.\label{eq:var_bound_4}
    %\end{align}
%Combining \eqref{eq:var_bound_3} and \eqref{eq:var_bound_4}, we obtain the desired result.  
Substituting \eqref{eq:var_bound_4} into \eqref{eq:var_bound_3}, the factor $m$ cancels with the prefactor $\frac{1}{mT}$, so the noise contributes
$\frac{1}{T\varepsilon^2}\, 2\ln(1.25/\delta)\big((N\pi_{\min})^{-1}|\lambda| + \ln 2\big)^2$,
which gives the desired result.
\qed

\subsection{Proof of \Cref{th:opt_lambda_no_noise}}
\begin{proof}
Using \Cref{eq:var_bound_2} in the proof of \Cref{th:gen_estimator_variance}, we have $\operatorname{Var} \left[ \hat{D}_{\operatorname{KL}} \left( \Pi \| P \right)_{[\lambda, T, \varepsilon, \delta]} \right]=$
    \begin{align}
        &\frac{1}{mT}\left(\lambda^{2} \mathbb{E}[r(X)^2] -\lambda^2 + \mathbb{E}[\ln^2{r(X)}] -D_{\operatorname{KL}}(\Pi\|P)^2\right.\nonumber\\
        &\left.- 2\lambda\left(D_{\operatorname{KL}}(P \| \Pi)+ D_{\operatorname{KL}}(\Pi \| P)\right)\right) +\frac{1}{mT}\operatorname{Var}\left[\eta_{\varepsilon,\delta}\right].\label{eq:opt_lambda_1}
    \end{align}
    Recalling that the setting of this theorem assumes no \ac{DP} noise has been added to the estimator, implying that $\eta_{\varepsilon,\delta}=0$, we essentially have  $\operatorname{Var}[\eta_{\varepsilon,\delta}]=0$. Hence, for fixed $P$, $\Pi$, $m$, and $T$, we note that \Cref{eq:opt_lambda_1} is a differentiable function w.r.t. $\lambda$. Accordingly, we write $H:\mathbb{R}\mapsto \mathbb{R}$ as the corresponding auxiliary function w.r.t. $\lambda$ s.t. 
    \begin{align}
        &H(\lambda)=\frac{1}{mT}\left(\lambda^{2} \mathbb{E}[r(X)^2] -\lambda^2 + \mathbb{E}[\ln^2{r(X)}] \right.\nonumber\\
        &\left.-D_{\operatorname{KL}}(\Pi\|P)^2  - 2\lambda\left(D_{\operatorname{KL}}(P \| \Pi)+ D_{\operatorname{KL}}(\Pi \| P)\right)\right).\nonumber
    \end{align}
    Recalling \Cref{lem:estimator_fraction_expect}, which essentially ensures that $\mathbb{E}[r(X)^2]-1$ equals $\operatorname{Var}[r(X)]$, we crucially observe that $\mathbb{E}[r(X)^2]\geq 1$, and thus conclude that $H(\lambda)$ is convex w.r.t. $\lambda$.

    Therefore, to find the optimal $\lambda_0$, it suffices to find the solutions in $\lambda=\lambda_0$ such that $\frac{dH}{d\lambda}(\lambda_0)=0$. For simplicity, let us write $\zeta_{\Pi}(X)=\mathbb{E}_{X\sim \Pi}[r(X)^2]-1$. We start by computing 
    \begin{align}
        &\frac{d H}{d \lambda}=\frac{1}{mT}\left(2\lambda \zeta_{\Pi}(X) -2 \left(D_{\operatorname{KL}}(P \| \Pi)+ D_{\operatorname{KL}}(\Pi \| P)\right)\right).\label{eq:opt_lambda_2}
    \end{align}
    Now we equate \Cref{eq:opt_lambda_2} with $0$ to get %$\lambda_0=\frac{D_{\operatorname{KL}}(P \| \Pi)+ D_{\operatorname{KL}}(\Pi \| P)}{\zeta_{\Pi}(X)}$.
    \begin{align}
      &\lambda_0=\frac{D_{\operatorname{KL}}(P \| \Pi)+ D_{\operatorname{KL}}(\Pi \| P)}{\zeta_{\Pi}(X)}.\label{eq:opt_lambda_3}
    \end{align}
     Next, observing that $\zeta_{\Pi}(X)=\operatorname{Var}_{X\sim \Pi}\left[r(X)\right]$, we get the desired result.% and recalling the definition of $\alpha$ as introduced in \Cref{th:gen_estimator_variance}, we ensure that $\frac{1}{\beta} -1\leq\zeta_{\Pi}(X)\leq \alpha-1$. Substituting these bounds in the value of $\lambda_0$ as derived in \Cref{eq:opt_lambda_3}, we get the desired result.
\end{proof}

\subsection{Proof of \Cref{corr:opt_lambda_noise}}
\begin{proof}
Let $\tilde{D}=D_{\mathrm{KL}}(P\|\Pi)+D_{\mathrm{KL}}(\Pi\|P)\ge 0$. By Theorem~V.1, $\lambda_0=\tilde{D}/\mathrm{Var}[r(X)]$, and by Fact~IV.1, $\mathrm{Var}[r(X)]=\mathbb{E}[r(X)^2]-1$. As in the proof of Theorem~IV.8,
\[
\mathbb{E}\big[r(X)^2\big]=\sum_{x\in\mathcal{X}}P(x)\,\frac{P(x)}{\Pi(x)}\le\alpha\sum_{x\in\mathcal{X}}P(x)=\alpha ,
\]
so $\mathrm{Var}[r(X)]\le\alpha-1$. Moreover, $\mathrm{Var}[r(X)]>0$: otherwise $r\equiv\mathbb{E}[r(X)]=1$ on the support of $\Pi$, \ie\ $P=\Pi$. Hence $0<\mathrm{Var}[r(X)]\le\alpha-1$, and
\[
\lambda_0=\frac{\tilde{D}}{\mathrm{Var}[r(X)]}\ \ge\ \frac{\tilde{D}}{\alpha-1}.
\]
\end{proof}

\subsection{Proof of \Cref{thm:opt-var}}
\begin{proof}
 Shadowing the proof of \Cref{th:opt_lambda_no_noise}, we use \Cref{eq:opt_lambda_1} to define an auxiliary function
$G:\mathbb{R}\to\mathbb{R}$ in terms of $\lambda$, which incorporates the noise term
$\eta_{\varepsilon,\delta}$. Writing $\tilde{D}=D_{\mathrm{KL}}(P\|\Pi)+D_{\mathrm{KL}}(\Pi\|P)\ge 0$,
\begin{align*}
G(\lambda)={}&\frac{1}{mT}\Big(\lambda^2\,\mathrm{Var}[r(X)]+\mathbb{E}[\ln^2 r(X)]\\
&-D_{\mathrm{KL}}(\Pi\|P)^2-2\lambda\tilde{D}\Big)+\frac{1}{mT}\mathrm{Var}[\eta_{\varepsilon,\delta}].
\end{align*}
By the definition of $\eta_{\varepsilon,\delta}$ in Algorithm~2,
\[
\frac{1}{mT}\mathrm{Var}[\eta_{\varepsilon,\delta}]
=\frac{2\ln(1.25/\delta)}{T\varepsilon^2}\Big(\frac{|\lambda|}{N\pi_{\min}}+\ln 2\Big)^2 .
\]
$G$ is continuous on $\mathbb{R}$, and convex and differentiable on each of $[0,\infty)$ and
$(-\infty,0)$. We minimise it on each half-line.

\emph{Case 1.} $\lambda\ge 0$: here $|\lambda|=\lambda$ and
\begin{align*}
\frac{dG}{d\lambda}={}&\frac{2}{mT}\Big(\lambda\,\mathrm{Var}[r(X)]-\tilde{D}\Big)\\
&+\frac{4\ln(1.25/\delta)}{T\varepsilon^2}\Big(\frac{\lambda}{N\pi_{\min}}+\ln 2\Big)\frac{1}{N\pi_{\min}} .
\end{align*}
Equating this with $0$ gives
\[
\lambda\Big(\mathrm{Var}[r(X)]+\frac{2m\ln(1.25/\delta)}{(\varepsilon N\pi_{\min})^2}\Big)
=\tilde{D}-\frac{m\ln(1.25/\delta)\ln 4}{\varepsilon^2N\pi_{\min}},
\]
that is, $\lambda=\lambda^\star$. If $\lambda^\star\ge 0$, it lies in $[0,\infty)$ and, by convexity,
minimises $G$ there. Otherwise $\frac{dG}{d\lambda}>0$ on $[0,\infty)$, so the minimum over
$[0,\infty)$ is attained at $\lambda=0$.

\emph{Case 2.} $\lambda<0$: here $|\lambda|=-\lambda$, so the noise term is
$\frac{2\ln(1.25/\delta)}{T\varepsilon^2}\big(\ln 2-\frac{\lambda}{N\pi_{\min}}\big)^2$ and
\begin{align*}
\frac{dG}{d\lambda}={}&\frac{2}{mT}\Big(\lambda\,\mathrm{Var}[r(X)]-\tilde{D}\Big)\\
&-\frac{4\ln(1.25/\delta)}{T\varepsilon^2}\Big(\ln 2-\frac{\lambda}{N\pi_{\min}}\Big)\frac{1}{N\pi_{\min}}<0,
\end{align*}
as the first term is non-positive and the second is negative. Hence $G$ is strictly decreasing on
$(-\infty,0)$ and, being continuous at $0$, satisfies $G(\lambda)>G(0)$ for all $\lambda<0$.

Combining the two cases, the minimiser of $G$ over $\mathbb{R}$ is $\lambda_0=\max\{0,\lambda^\star\}$.
\end{proof}

\subsection{Proof of \Cref{corr:opt_lambda_order}}
\begin{proof}

Let $\tilde{D}(\Pi,P)=D_{\mathrm{KL}}(P\|\Pi)+D_{\mathrm{KL}}(\Pi\|P)$ and
$\omega=\frac{m\ln(1.25/\delta)\ln 4}{\varepsilon^2N\pi_{\min}}$. Then, using \Cref{th:opt_lambda_no_noise}, the optimal
variance parameter for \textsc{FedPriKL} without DP noise reduces down to
$\frac{\tilde{D}(\Pi,P)}{\zeta_\Pi(X)}$, where $\zeta_\Pi(X)$ is as defined in the proof of
\Cref{th:opt_lambda_no_noise}. By \Cref{thm:opt-var}, the optimal parameter with DP noise is
$\lambda_0=\max\big\{0,\frac{\tilde{D}(\Pi,P)-\omega}{\zeta_\Pi(X)+\omega\ell}\big\}\ge 0$ with
$\ell=(N\pi_{\min}\ln 2)^{-1}$, so $|\lambda_0|=\lambda_0$. If $\tilde{D}(\Pi,P)>\omega$, comparing
the numerators and the denominators gives
$\frac{\tilde{D}(\Pi,P)-\omega}{\zeta_\Pi(X)+\omega\ell}<\frac{\tilde{D}(\Pi,P)}{\zeta_\Pi(X)}$.
Otherwise $\lambda_0=0<\frac{\tilde{D}(\Pi,P)}{\zeta_\Pi(X)}$ whenever $P\neq\Pi$. In both cases
the desired result follows.
\end{proof}

\end{document}